\documentclass{article}

\usepackage[preprint]{paper}

\usepackage{booktabs}
\usepackage{multirow}
\usepackage{makecell}
\usepackage{xcolor}
\usepackage{colortbl}
\usepackage{graphicx}
\usepackage{subcaption}
\usepackage{amsmath}
\usepackage{amssymb}
\usepackage{mathtools}
\usepackage{url}
\usepackage{enumitem}
\usepackage{array}
\usepackage{adjustbox}
\usepackage{wrapfig}
\usepackage{microtype}
\usepackage{longtable}
\usepackage{caption}
\usepackage{comment}

\definecolor{l0col}{RGB}{60,140,60}
\definecolor{l1col}{RGB}{50,100,180}
\definecolor{l2col}{RGB}{185,135,30}
\definecolor{l3col}{RGB}{175,55,50}
\definecolor{cellbest}{RGB}{198,232,198}
\definecolor{cellsec}{RGB}{228,244,228}

\title{\includegraphics[height=1.2cm]{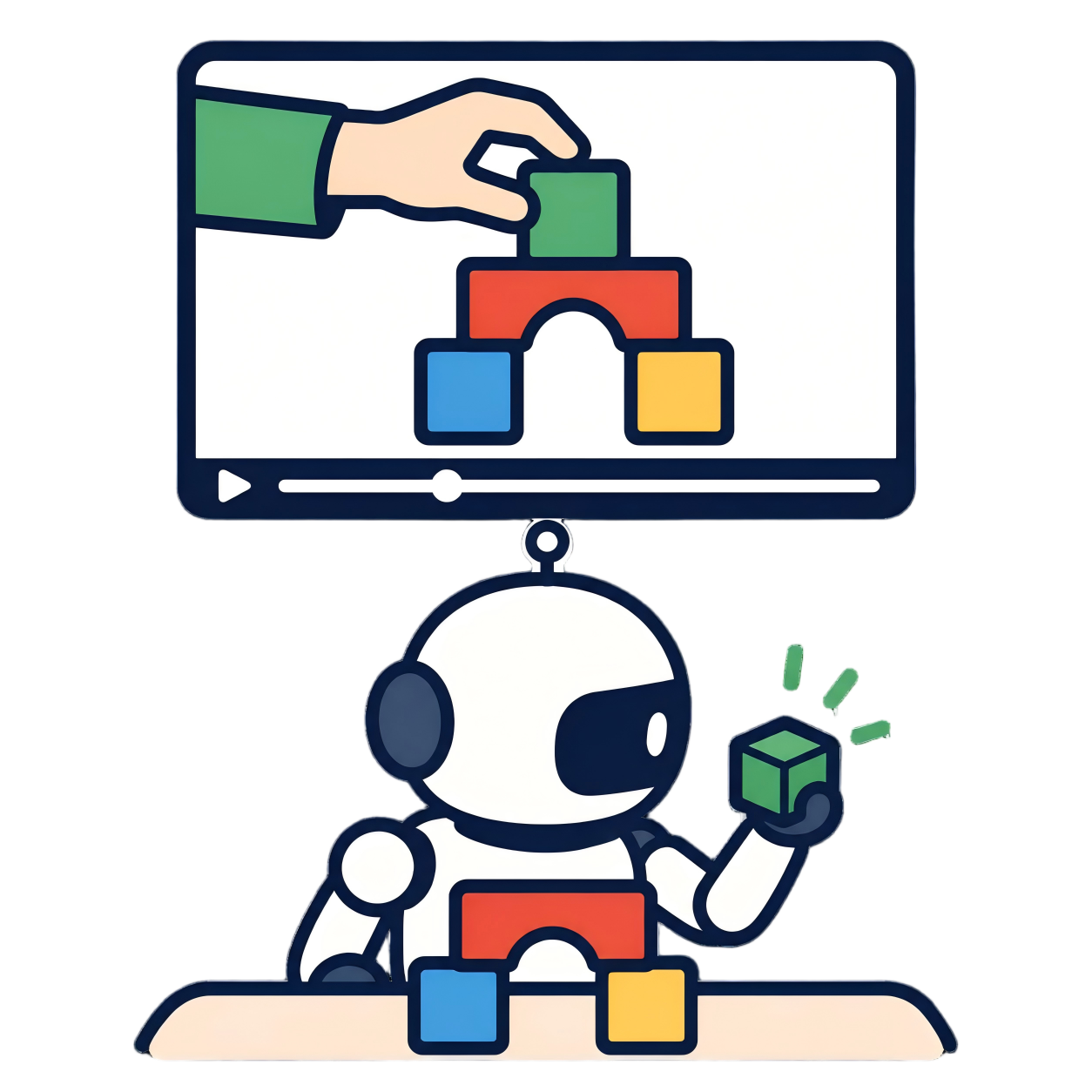} The Imitator Game: Benchmarking Robot Imitative Ability Beyond Action Prediction}

\author{
  \textbf{Xunzhe Zhou}$^{1,2,\ast,\dagger}$ \quad
  \textbf{Yiyang Cai}$^{2,3,\ast}$ \quad
  \textbf{Fengyi Wang}$^{2,3,\ast}$ \quad
  \textbf{Ran Ju}$^{1,2,\ast}$ \quad
  \textbf{Hanxiang Ren}$^{2,4}$ \quad \\[1mm]
  \textbf{Ruizhe Liu}$^{1}$ \quad
  \textbf{Yu Zhang}$^{1}$ \quad
  \textbf{Qian Luo}$^{1,2}$ \quad
  \textbf{Feng Chen}$^{1}$ \quad
  \textbf{Pei Zhou}$^{1,2}$ \quad \\[1mm]
  \textbf{Yi Ma}$^{1,2}$ \quad
  \textbf{Yanchao Yang}$^{1,2,\ddagger}$ \\[2mm]
  {\normalsize $^{1}$The University of Hong Kong \quad $^{2}$TranscEngram \quad $^{3}$Fudan University \quad $^{4}$Zhejiang University} \\[2mm]
  {\normalsize $^{\ast}$Equal contribution \quad $^{\dagger}$Project lead \quad $^{\ddagger}$Corresponding author} \\[4mm]
  {\normalsize \texttt{\url{https://imitator-game.github.io}}} \\[-4mm]
}

\begin{document}
\maketitle

\begin{figure}[th]
  \centering
  \includegraphics[width=\textwidth]{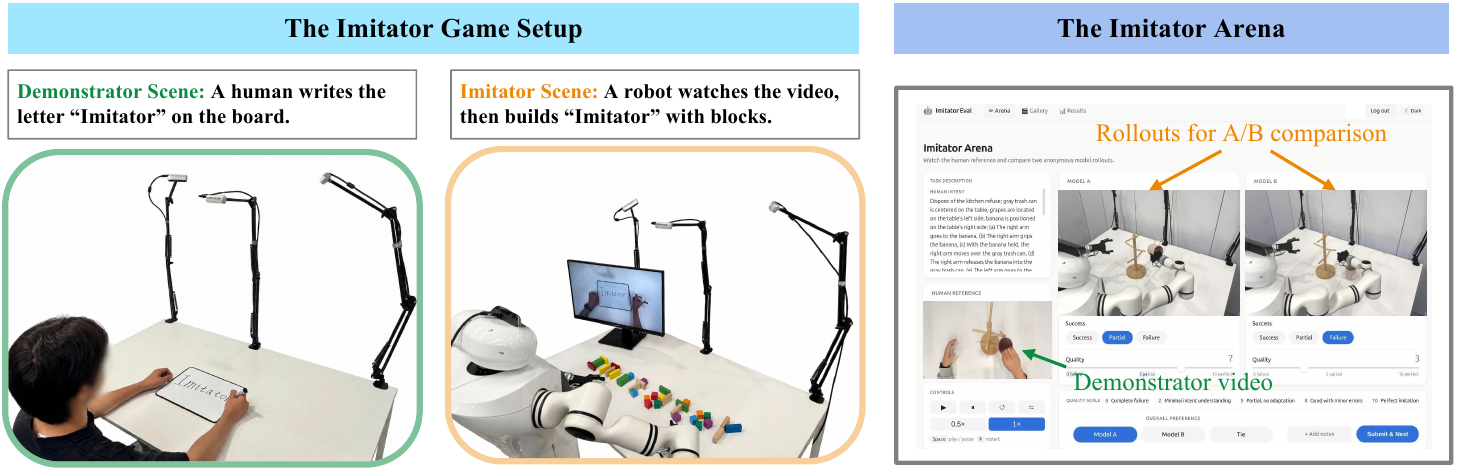}
  \caption{\textbf{The Imitator Game.} \textit{(Left)} The Imitator Game setup: a human demonstrates a manipulation task in the Demonstrator Scene (green), and a robot watches the video and reproduces the underlying \emph{intent} in the Imitator Scene (orange) using whatever objects are available.
  \textit{(Right)} The Imitator Arena evaluation platform: human evaluators can make structured blind A/B comparisons of how two anonymized model rollouts imitate against the demonstrator video.}
  \label{fig:teaser}
\end{figure}

\begin{abstract}
Humans imitate at the level of intent: given a demonstration, we infer its goal and carry it out with whatever tools, objects, and layouts are at hand. Current robot policies instead learn observation-to-action mappings from visual inputs and language instructions, without explicitly inferring the demonstrated task. Learning from human video thus remains largely trajectory-level: models can replay motions in near-identical scenes, but still struggle to imitate what the demonstrator intends rather than merely what they do. We introduce \textbf{The Imitator Game}, a four-level benchmark (L0--L3) that progressively widens the gap between the human demonstration and the robot's own scene, isolating where trajectory replay ceases to suffice and task understanding becomes necessary. We pair it with \textbf{IG-10K}, the largest environment-aligned paired human--robot dataset to date and the only one instantiated across all four levels in both real and simulated settings (20,000+ paired episodes, 50+ tasks, 6 domains), and \textbf{Imitator Arena}, an open platform for blind A/B human evaluation. Across nine state-of-the-art models, performance is stable from L0 to L2 but collapses at L3, identifying functional substitution -- achieving the same intent through a different object affordance -- as the decisive barrier to intent-level imitation. Human-video-conditioned models outperform caption-conditioned ones, yet every model falls below $13\%$ zero-shot success on unseen tasks; fine-tuning IG-10K-pretrained models with only $10$ paired human-robot demonstrations yields large gains that grow with pretraining scale.
\end{abstract}

\section{Introduction}
\label{sec:intro}

When people watch someone stir soup, they do not memorize a sequence of motions. They infer a goal-directed behavior: the soup needs to be stirred, with whatever is available. If the pot is moved, the spoon replaced with chopsticks, or the bowl exchanged for a wok, people adapt immediately while preserving the underlying intent. This form of \emph{goal-directed imitation}~\citep{tomasello2009cultural} -- reproducing the purpose of an action rather than its exact motion -- is a hallmark of human and great-ape cognition.

Robots do not yet share this ability. Most approaches for learning from human video rely on \emph{trajectory remapping}: extracting hand keypoints or optical flow from a demonstration~\citep{qin2022dexmv,xu2023xskill}, retargeting the motion to robot actions, and replaying it in a closely matched scene~\citep{sivakumar2022robotic,lepert2025masquerade}. These methods are useful for bootstrapping robot behaviour, but they depend heavily on the demonstration and execution scenes remaining similar. Small changes in object placement, geometry, or available tools can already invalidate the demonstrated motion, even when the underlying task itself is unchanged.

The question we ask differs from that of prior work. Existing generalization benchmarks~\citep{pumacay2024colosseum,sedlacek2026realm} mainly evaluate robustness to perturbations of scenes encountered during training, while video-conditioned policies~\citep{jain2024vid2robot,kim2025uniskill,tang2025trajectory} are typically tested in-distribution, with only limited mismatch between the demonstration and execution environments. In contrast, we evaluate \emph{imitative ability}: whether a policy can watch a human demonstration video and reproduce its intent in a previously unseen scene, even when the available objects differ substantially in placement, geometry, or function from those originally shown in the demonstration. The central challenge is not trajectory reproduction itself, but determining what aspects of the demonstrated behavior should still be preserved under a changed scene. This ability is ultimately what determines whether the large amount of human video already available online can serve as a practical supervision source for robot manipulation.

Drawing on work in cultural learning~\citep{tomasello2009cultural} and goal-directed imitation~\citep{meltzoff1995understanding}, we formulate this ability as four levels of increasing mismatch between the demonstration and the robot's own scene:

\begin{itemize}[leftmargin=2em, itemsep=1pt, topsep=2pt, parsep=0pt]
\item[\textcolor{l0col}{\textbf{L0}}] \textbf{Scene-identical execution.}
The imitator scene matches the demonstration scene; the policy is expected to reproduce the demonstrated trajectory.

\item[\textcolor{l1col}{\textbf{L1}}] \textbf{Spatial adaptation.}
Object identities are preserved but their spatial configuration differs; the policy must achieve the same object-level outcome through a different trajectory.

\item[\textcolor{l2col}{\textbf{L2}}] \textbf{Visual-physical generalization.}
Task-relevant objects differ in appearance or geometry while preserving semantic role; the policy must generalize across visually or physically different instances of the same object category.

\item[\textcolor{l3col}{\textbf{L3}}] \textbf{Intent-level transfer.}
The demonstrated object is replaced by one of different semantics or function (e.g., teapot $\to$ water tap for ``fill the cup''); the policy must infer the underlying intent and realize it through a different affordance structure.
\end{itemize}

\vspace{4pt}
\noindent \textbf{Contributions.}
We introduce \textbf{the Imitator Benchmark}, a four-level evaluation framework for robot video imitation that measures imitative capability under progressively increasing mismatch between demonstration and execution scenes. To support this benchmark, we construct \textbf{IG-10K}, the largest environment-aligned human--robot paired manipulation dataset to date, containing 20{,}000+ paired episodes, 200+ task variants, and 6 domains across both simulation and the real world. We further introduce \textbf{Imitator Arena}, an open platform for structured blind human evaluation of robot imitation behavior, and perform a systematic evaluation of 9 state-of-the-art models (15 trained variants spanning language-conditioned VLAs, cross-embodiment skill methods, and video-conditioned visuomotor policies) across multiple training scales and regimes.

\section{Related Works}

\paragraph{Learning manipulation from human video.}
Prior work learns from human videos by retargeting extracted hand or object poses~\citep{qin2022dexmv,sivakumar2022robotic}, following optical-flow or object-centric motion cues~\citep{tang2025trajectory,dan2025x}, inpainting human hands as robot grippers~\citep{lepert2025masquerade}, or filtering cross-embodiment actions~\citep{dan2025x}. Skill-representation methods such as XSkill~\citep{xu2023xskill}, UniSkill~\citep{kim2025uniskill}, RHyME~\citep{kedia2025one}, and ImMimic~\citep{liu2025immimic} learn human-robot correspondences, and Vid2Robot~\citep{jain2024vid2robot} conditions control on a video prompt. However, these methods assume the robot scene is largely congruent with the demonstration and are evaluated in-distribution, without graded mismatch between the two.

\paragraph{Benchmarks and datasets for manipulation.}
Existing benchmarks study robustness, long-horizon reasoning, and generalization in robot manipulation. Benchmarks such as COLOSSEUM~\citep{pumacay2024colosseum}, RoboTwin 2.0~\citep{chen2025robotwin}, and REALM~\citep{sedlacek2026realm} perturb visual, physical, embodiment, or environment factors, while CALVIN~\citep{mees2022calvin}, LIBERO~\citep{liu2023libero}, and VLABench~\citep{zhang2025vlabench} focus on compositional language-conditioned tasks. Large-scale datasets including Open X-Embodiment~\citep{o2024open}, DROID~\citep{khazatsky2024droid}, BridgeData V2~\citep{walke2023bridgedata}, and FMB~\citep{luo2025fmb} instead emphasise diversity across tasks, scenes, and embodiments. In contrast, IG-10K pairs human demonstrations with progressively mismatched robot scenes, allowing direct evaluation of transfer from human video and intent-level imitation.

\paragraph{Robot manipulation models.}
Robot manipulation policies range from specialised visuomotor methods -- such as ACT~\citep{zhao2023learning}, Diffusion Policy~\citep{chi2025diffusion}, and VQ-BeT~\citep{lee2024behavior} -- to large-scale VLA and diffusion-based foundation models including OpenVLA~\citep{kim2024openvla,kim2025fine}, $\pi_{0.5}$~\citep{black2024pi_0,intelligence2025pi_}, the GR00T series~\citep{bjorck2025gr00t}, RDT-1B~\citep{liu2025rdt}, and H-RDT~\citep{bi2026h}. These systems have advanced language conditioning, action generation, cross-robot pretraining, and few-shot adaptation, but are typically evaluated on executing a specified task within the robot's own scene. In this work, we instead study whether a policy can infer and reproduce the demonstrated behaviour from a human video under mismatched execution conditions, where the robot scene may differ substantially from the demonstration itself.

\section{The Imitator Game}
\label{sec:method}

This section has two parts. Section~\ref{sec:capability} formalizes the \emph{imitator capability} -- the ability to infer and reproduce the behavior demonstrated in a human video -- as a four-level hierarchy, where each level corresponds to a different degree of scene mismatch and the corresponding fidelity of imitation. Section~\ref{sec:toolkit} introduces the \emph{Imitator Benchmark}: \textbf{IG-10K}, the largest environment-aligned human--robot paired dataset to date; \textbf{Imitator Arena}, a unified platform for simulation and real-world evaluation with both automated metrics and a human-evaluation interface; and the adaptation of three families of state-of-the-art baselines, each trained on IG-10K and evaluated in the Arena.

\subsection{The Imitator Capability}
\label{sec:capability}

\subsubsection{Problem setting and scene decomposition}
\label{sec:problem-setting}

\textbf{Problem setting.}
The imitator is given a human demonstration video $V$, recorded in some scene $\mathcal{S}_{\text{demo}}$. The task is defined implicitly by whatever is demonstrated in $V$ -- there is no separate task description, goal predicate, or symbolic specification. The imitator is then placed in its own scene $\mathcal{S}_{\text{imit}}$, where it receives observations $o_t$ and must generate actions $a_t$ that reproduce the demonstrated task as faithfully as $\mathcal{S}_{\text{imit}}$ permits. Formally, an imitator policy is a mapping $\pi\bigl(a_t \mid o_t, V\bigr)$; goal predicates are used only inside the evaluation platform for simulation scoring, never as policy input.

\textbf{Scene decomposition.}
We represent a scene as an object collection and its spatial configuration,
\(
\mathcal{S} = \bigl(\{O_i=(A_i, G_i, S_i)\}, P\bigr),
\)
where $A_i$, $G_i$, and $S_i$ denote the appearance, geometry, and semantic category of object $O_i$, respectively, and $P$ denotes the spatial configuration of the object collection. In our specific definition context, semantic categories are distinguished mainly by the general purposes of objects, and spatial configuration primarily concerns the initial placement of task-relevant objects at the start of the episode.
The distinction between intrinsic object properties and extrinsic configuration is central to the hierarchy: changing $P$ preserves object identity but changes where objects are placed, whereas changing $(A,G,S)$ changes the behavior logic to accomplish the tasks shown in the demonstration scene. For the functional task objects in IG-10K, we use the operational assumption
\[
S \text{ changes} \;\Longrightarrow\; G \text{ changes} \;\Longrightarrow\; A \text{ changes},
\]
while the reverse directions do not generally hold. This assumption is not intended as a general ontology of objects; it is used to construct benchmark instances where each level transition is induced by one interpretable mismatch. It also implies that comparing $P$ is meaningful only when demonstration and imitator scenes share a common object set. 
Edge cases are discussed in Appendix~\ref{app:scene-decomposition}.

\subsubsection{Four levels of imitation}
\label{sec:levels}

\begin{figure}[t]
\centering
\includegraphics[width=\textwidth]{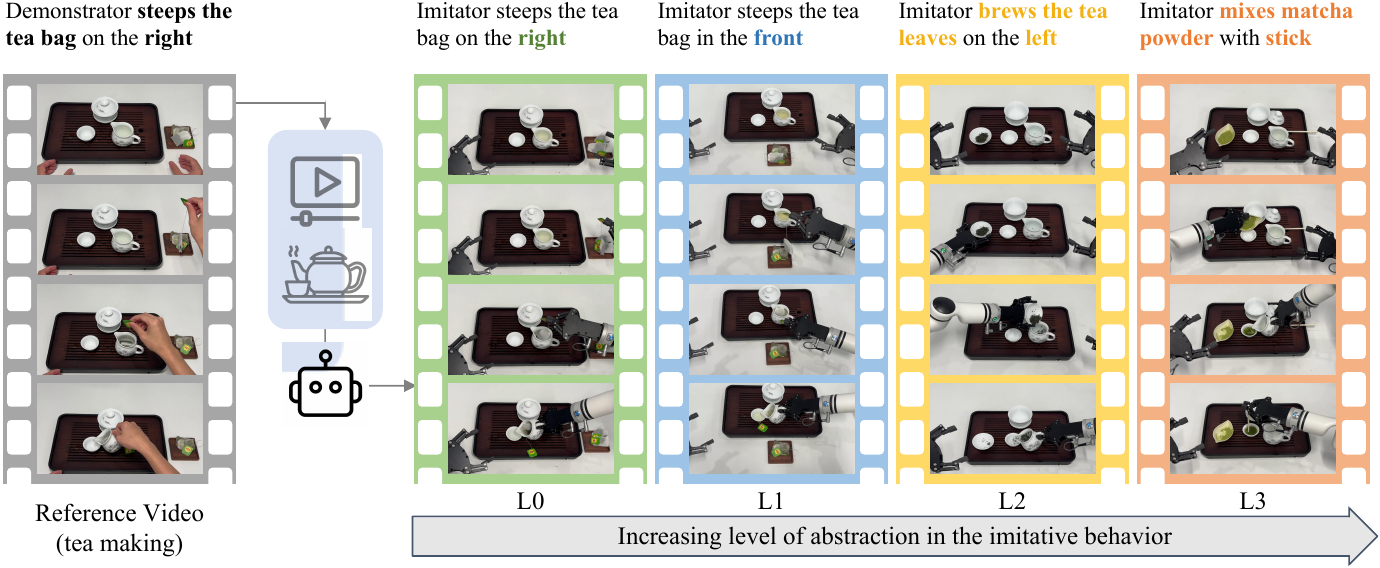}
\caption{\textbf{Four levels of imitation,} illustrated on a single demonstration: a human \emph{making tea by steeping a tea bag} (leftmost filmstrip). Left to right, the imitator's scene diverges further from the demonstration, so less and less of the demonstration can still be copied exactly. \textbf{L0:} same objects and layout; the imitator steeps the tea bag on the right, reproducing the demonstrated motion. \textbf{L1:} the layout is rearranged (tea bag in front rather than on the right), so the same final state must be reached by a different trajectory. \textbf{L2:} the tea bag is replaced by loose tea leaves of the same kind, so only the semantic task ``make tea'' is reproducible. \textbf{L3:} no tea is present, so the imitator serves the same intent with the available matcha powder, mixing it rather than steeping. Abstraction increases left to right.}
\label{fig:taxonomy}
\end{figure}

\begin{table}[th]
\centering
\small
\caption{\textbf{The four levels of the imitator game} as preservation patterns across three scene-property columns. \checkmark\ = preserved between $\mathcal{S}_{\text{demo}}$ and $\mathcal{S}_{\text{imit}}$; $\times$ = changed; n/a = not applicable, as the object set has changed and configuration has no identity mapping. The right column states the highest fidelity the imitator's scene permits.}
\label{tab:levels}
\begin{tabular}{lcccl}
\toprule
Level & $P$ & $A$ or $G$ & $S$ & Replication fidelity expected of the imitator \\
\midrule
L0 (trajectory)         & \checkmark & \checkmark & \checkmark & Trajectory match: similar motion to $V$. \\
L1 (object end-state)   & $\times$   & \checkmark & \checkmark & Same final object states as $V$; trajectory may differ. \\
L2 (semantic task)      & n/a        & $\times$   & \checkmark & Same task semantics as $V$; final states differ. \\
L3 (affordance-adapted) & n/a        & $\times$   & $\times$   & Same underlying intent, via re-purposed affordances. \\
\bottomrule
\end{tabular}
\end{table}

How closely the imitator can copy the demonstration is decided by its own scene $\mathcal{S}_{\mathrm{imit}}$. As $\mathcal{S}_{\mathrm{imit}}$ diverges away from $\mathcal{S}_{\mathrm{demo}}$ along $(P,A,G,S)$, what can still be matched becomes less and less specific: first the exact motion, then the exact final state of each object, then only the task in the ordinary sense of the word, and finally only the purpose behind it. Table~\ref{tab:levels} defines this hierarchy as preservation patterns over $P$, $A\text{-or-}G$, and $S$; Figure~\ref{fig:taxonomy} illustrates the progression on a single demonstration. We treat $P$ as preserved when the motion shown in $V$ would still work if it were replayed directly in $\mathcal{S}_{\mathrm{imit}}$, and as changed otherwise. The $A\text{-or-}G$ column is preserved only when the task-relevant object set is intrinsically identical, and $S$ is preserved when each task-relevant object keeps its semantic category. Thus, each $\mathrm{L}{\to}\mathrm{L}{+}1$ transition permits greater differences between the demonstration and the imitator scene, while correspondingly broadening the criterion for successful imitation. This gives the hierarchy a clear progression from matching actions, to reproducing object outcomes, to preserving semantic function, and ultimately to fulfilling the underlying intent.
Additional examples and edge cases are deferred to Appendix~\ref{app:levels}.

\subsection{The Imitator Benchmark}
\label{sec:toolkit}

The benchmark comprises three reusable artifacts. \textbf{IG-10K} (Section~\ref{sec:dataset}) is the largest environment-aligned human-robot paired manipulation dataset to date, instantiated across all four levels of Table~\ref{tab:levels}. \textbf{Imitator Arena} (Section~\ref{sec:arena}) supports automated rule-based scoring in simulation and structured human evaluation. \textbf{Baseline adapters} (Section~\ref{sec:baselines}) let three different families of methods share the same input with human demonstration videos and robot observation. Using these we train or fine-tune nine state-of-the-art models, yielding fifteen trained variants, evaluated uniformly within the Arena.

\subsubsection{IG-10K: the largest environment-aligned paired human-robot manipulation dataset}
\label{sec:dataset}

\begin{figure}[t]
\centering
\includegraphics[width=\textwidth]{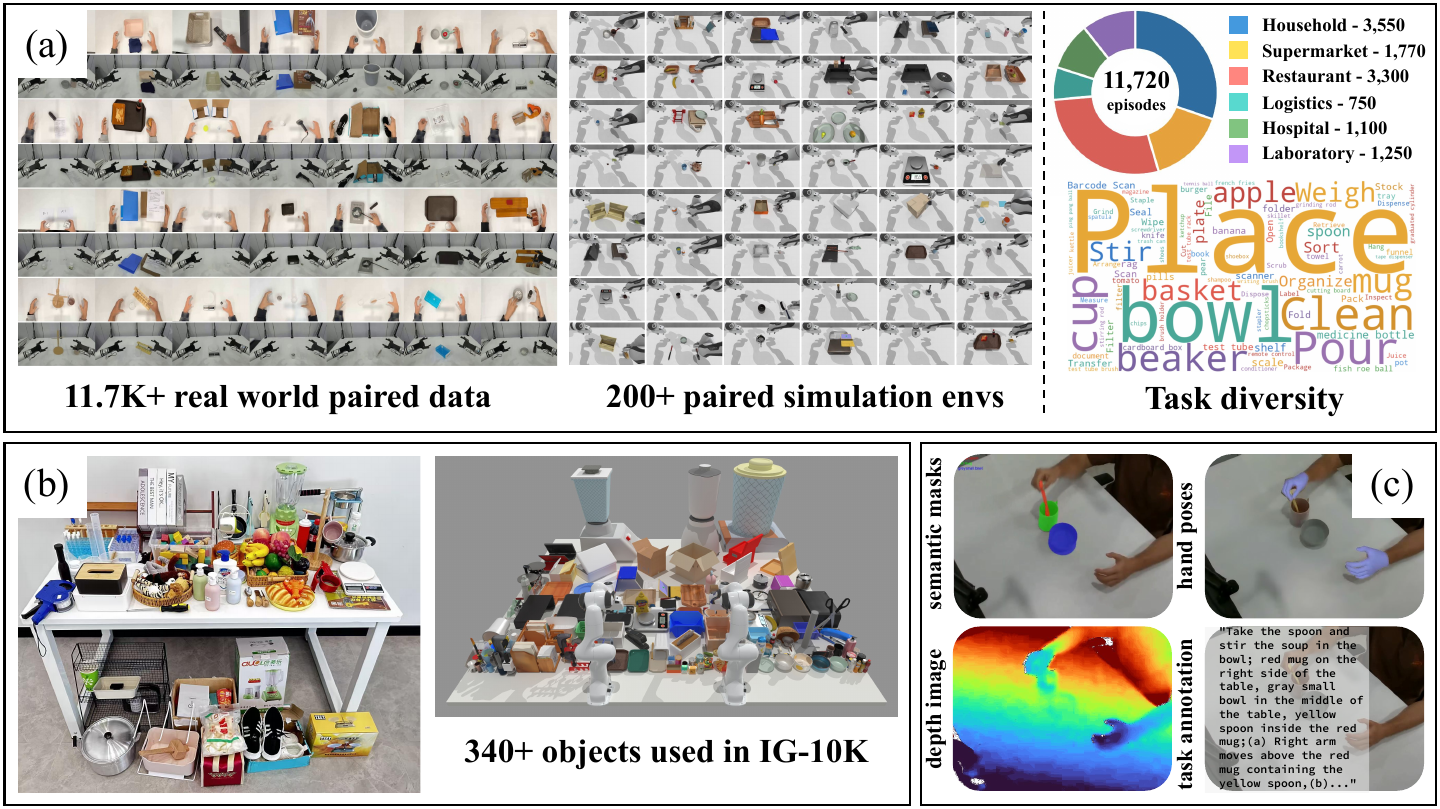}
\caption{\textbf{The IG-10K dataset.} \emph{(a)} IG-10K contains $20{,}000+$ paired episodes: 11.7K real human-robot paired episodes, each with one human clip and one corresponding real-robot clip, plus 10K paired simulation episodes. These span household, supermarket, restaurant, logistics, hospital, and laboratory manipulation tasks. The montage, donut chart, and word cloud summarize the dataset scale and task diversity. \emph{(b)} The object gallery shows the real and simulated assets used to instantiate the L0--L3 scene variations. \emph{(c)} Each episode is released with dense offline annotations, including semantic masks, hand poses, depth images, and language descriptions.}
\label{fig:dataset}
\end{figure}

\paragraph{Paired-episode design.}
The atomic unit of IG-10K is a \emph{paired episode}: a human demonstration video $V$ in $\mathcal{S}_{\text{demo}}$, together with one or more robot episodes in scenes $\mathcal{S}_{\text{imit}}$ whose relation to $\mathcal{S}_{\text{demo}}$ instantiates one row of Table~\ref{tab:levels} -- from a shared object set and configuration (L0), through rearrangement (L1) and same-category substitution (L2), intent-base functional substitution (L3). The levels are therefore decided by the demonstration-execution \emph{pairs}, which is why the Arena supports fair level-agnostic comparison across anonymous policies for human evaluators (Section~\ref{sec:arena}). 
Hence, level is not a property of a task: the same human clip can serve four levels of its task, and only the robot-side scene changes. At this stage, each task-level pair uses exactly one substitution pattern for equal supervision across levels. More variation is left to future versions of the benchmark.

\paragraph{Task suite and scale.}
IG-10K contains 50+ base tasks across six domains -- Household (12), Supermarket (7), Restaurant (17), Logistics (4), Hospital (6), Laboratory (7), yielding 200+ task variants with four levels each suite, 50+ paired episodes each. Since the number of possible substitutions grows as the level rises with more mismatch, each task is paired with one substitution pattern per level at this stage for comparable comparison in levels. More variation is left to future versions of the benchmark. A few tasks omit L3 where substitution would be ill-defined or unsafe, e.g.\ hazardous-liquid handling in the laboratory. In total IG-10K provides 20{,}000+ paired episodes, released in \texttt{lerobot-0.5.0} format.
Each episode is multi-view and richly annotated -- 3D MANO hand pose, segmentation masks, and multi-abstraction language -- placing IG-10K ahead of prior resources on the axes that matter for imitation (Table~\ref{tab:datasets}). Real data are teleoperated in VR on a dual-arm Realman platform, and simulated data on a dual-arm Franka in ManiSkill3 with OMPL-planned trajectories over hand-crafted waypoints; the two are kept as separate domains throughout, but share the data format, the human-to-robot pairing, and the task split. Simulated pairs reuse the \emph{real} human clip of the same task and replace only the imitator scene; MANO hand pose is kept only for real-world human episodes. Collection platforms, assets, the camera rig, the annotation toolchain, and representative episodes (Figure~\ref{fig:dataset} \& Figure~\ref{fig:platform}) are detailed in Appendix~\ref{app:dataset}.

\begin{table}[t]
\centering
\scriptsize
\setlength{\tabcolsep}{3pt}
\caption{\textbf{IG-10K in context} vs.\ existing human-robot paired manipulation datasets. \checkmark/$\times$ = present/absent; $\Delta$ = partial. \textbf{Env.\ aligned}: human and robot demos share the same or explicitly corresponding task scene, objects, and setting (not necessarily frame-level timing). \textbf{Sim.\ env.}: a paired, interactive simulation environment for the human demos (robot-rendered videos or offline trajectories do not count).}
\label{tab:datasets}
\resizebox{\textwidth}{!}{%
\begin{tabular}{lcccccccccc}
\toprule
\textbf{Dataset} &
\makecell{\textbf{Human}\\\textbf{clips}} &
\makecell{\textbf{Robot}\\\textbf{clips}} &
\makecell{\textbf{Tasks}} &
\makecell{\textbf{Depth}} &
\makecell{\textbf{Env.}\\\textbf{aligned}} &
\makecell{\textbf{Sim.}\\\textbf{env.}} &
\makecell{\textbf{Multi}\\\textbf{view}} &
\makecell{\textbf{Hand}\\\textbf{pose}} &
\makecell{\textbf{Lang.}\\\textbf{ann.}} &
\makecell{\textbf{Seg.}\\\textbf{mask}} \\
\midrule
MIME~\citep{sharma2018multiple}        & 8.3k  & 8.3k  & 20 task families      & \checkmark & \checkmark & $\times$   & \checkmark & $\times$   & $\times$   & $\times$ \\
BC-Z~\citep{jang2022bc}         & 18.7k & 25.9k & 100+                  & $\times$   & $\times$   & $\times$   & $\times$   & $\times$   & \checkmark & $\times$ \\
Vid2Robot~\citep{jain2024vid2robot} & 10k & 100k & \makecell{Unknown}   & $\times$   & $\Delta$   & $\times$   & $\times$   & $\times$   & $\times$   & $\times$ \\
RH20T~\citep{fang2023rh20t}      & 110k  & 110k  & 147                   & \checkmark & $\times$   & $\times$   & \checkmark & $\times$   & \checkmark & $\times$ \\
H\&R~\citep{xie2026human2robot} & 2.6k  & 2.6k  & 10                    & $\times$   & \checkmark & $\times$   & $\times$   & $\times$   & $\times$   & $\times$ \\
DexWild~\citep{tao2025dexwild}  & 9.3k  & 1.4k  & \makecell{5 task families} & $\times$ & $\times$ & $\times$   & $\times$   & \checkmark & $\times$   & $\times$ \\
HRDexDB~\citep{lim2026hrdexdb}  & 1.4k  & 1.4k  & \makecell{Grasping}   & $\times$   & \checkmark & $\times$   & \checkmark & \checkmark & $\times$   & $\times$ \\
\textbf{IG-10K (Ours)}       & \textbf{11.7k} & \textbf{11.7k} & \textbf{50 tasks with 4 levels} & \textbf{\checkmark} & \textbf{\checkmark} & \textbf{\checkmark} & \textbf{\checkmark} & \textbf{\checkmark} & \textbf{\checkmark} & \textbf{\checkmark} \\
\bottomrule
\end{tabular}
}
\end{table}

\subsubsection{Imitator Arena and metrics}
\label{sec:arena}

Imitator Arena evaluates any policy following the imitator-game paradigm through two complementary score streams: automated goal-centric metrics in simulation, where the scene state is fully observable, and structured human preference judgments in both simulation and the real world.

\paragraph{Unified interface.}
Every policy receives the same specification -- the demonstration video $V$ and the observation stream $\{o_t\}$ -- an generates $a_t = \pi(V, o_t)$. Video-conditioned policies take these videos and images as they are; language-conditioned VLAs reach the demonstration videos through a platform-side step in which the Arena applies a fixed, deterministic captioner $T(V)$ and supplies the result as language conditioning.
To make it a fair comparison between VLAs and video-conditioned models, we keep captions dense and as informative as the video itself -- task intent, object layout, and the per-arm sub-task sequence with its motions -- and verify them by hand; see details in Appendix~\ref{app:baselines}. $T$ is fixed across all VLA baselines, so comparisons can reflect VLA capacities rather than how $V$ is captioned.
Crucially, the episode \emph{level} is never communicated to the policy: like a person watching the video, the imitator is not told how far its own scene has been changed, and has to work out from $V$ and its observations how much of the demonstration is still reproducible.

\paragraph{Human evaluation.}
Human evaluation provides the level-sensitive scoring channel for both simulation and real-world rollouts. For each episode, evaluators view three synchronized videos: the demonstration $V$ and two anonymized, randomly ordered rollouts from different models on the same imitator scene. The episode level is withheld, so evaluators judge what fidelity the scene permits from the demonstration--rollout pair itself with subjective human preference in imitative capability rather than from an explicit difficulty label. They answer three questions, worded the same way at every level: per-rollout \emph{task completion} (\textsc{success}/\textsc{partial}/\textsc{fail}), per-rollout \emph{imitation quality} $q\in[0,10]$ on a fixed five-anchor rubric, and pairwise \emph{overall preference}. These judgments yield $\mathrm{SR}_{\mathrm{human}}$ (counting only \textsc{success}), mean imitation score $\overline{Q}$, and per-model win rate $\mathrm{WR}$.
Two designs guarantee the fairness and quality of human evaluation. First, an A/B screen pairs two rollouts only when they are directly comparable: rollouts are bucketed by transfer setting (seen, zero-shot, transfer) and, within a bucket, paired only when they share the same domain, task, level, and episode, so the two videos differ only in the policy. Second, each screen is answered by a single evaluator, which keeps judgments independent. Reliability comes from repetition in random evaluation pair rollouts.
The bucketing protocol and formal aggregation rules are given in Appendix~\ref{app:arena}.

\paragraph{Automated metrics.}
In simulation, two hand-crafted metrics provide fast, reproducible proxies for absolute success: the \emph{final success rate} $\mathrm{SR}$, the fraction of episodes in which all task-relevant goal-state predicates hold at termination, and the denser \emph{sub-goal success rate} $\mathrm{Sub\text{-}SR}$, the mean fraction of a task's ordered sub-goal phases completed. Both are level-agnostic, purely reflecting the task completion at the specific task and level. The comparison between each level is therefore fair since we didn't impose any hierarchical human priors for any level, and leaves the level-by-level view to the cross-level analysis. Formal definitions and per-task predicate construction are in Appendix~\ref{app:arena}.

\subsubsection{Baseline adapters}
\label{sec:baselines}

Three families of existing methods are evaluated under the same interface of Section~\ref{sec:arena}, all receiving the same inputs ($V$, $o_t$) and trained or fine-tuned exclusively on IG-10K (Figure~\ref{fig:baselines}). Across all families, the video encoder or vision-language backbone is frozen, and only the action-generation module is trained. This keeps the adaptation budget comparable across methods and isolates how different policy architectures use the same human-video specification. As a result, the benchmark evaluates differences in action-generation interfaces rather than differences in large-scale representation adaptation. Full backbone fine-tuning is outside the scope of this comparison, since it would require substantially different optimization recipes, memory budgets, and training strategies across VLA, skill-based, and visuomotor models. Freezing also trades some accuracy for comparability and cost. In the engineering implementation, we cache the backbone activations, achieving orders-of-magnitude faster training at our evaluation scale.
\textbf{VLA models} (GR00T-N1.6~\citep{bjorck2025gr00t}, RDT-1B~\citep{liu2025rdt}, $\pi_{0.5}$~\citep{intelligence2025pi_}, OpenVLA~\citep{kim2024openvla}) are language-conditioned, so the fixed captioner $T(V)$ of Section~\ref{sec:arena} supplies the prompt. \textbf{Skill-based models} (XSkill~\citep{xu2023xskill}, UniSkill~\citep{kim2025uniskill}) learn shared human-robot skill representations and act on the retrieved skill. \textbf{Vision-action models} (ACT~\citep{zhao2023learning}, Diffusion Policy~\citep{chi2025diffusion}, VQ-BeT~\citep{lee2024behavior}) gain video conditioning by concatenating a task embedding from one of three frozen encoders (DINOv2-ViT-L/14~\citep{oquab2023dinov2}, SigLIP2-SO400M~\citep{tschannen2025siglip}, VideoMAE-Large~\citep{tong2022videomae}) to the observation tokens. Per-family mechanisms, encoders, and recipes are in Appendix~\ref{app:baselines}.

\paragraph{The protocol does not favor any kind of model.} By spanning language-conditioned generation (VLA), cross-embodiment skill retrieval, and video-conditioned action regression (VA) through the same interface, on identical data and identical episodes, the Arena is insensitive to the architectural paradigm under test. Whether any current paradigm closes the gap between trajectory-level imitation at L0 and intent-level affordance adaptation at L3 is the empirical question taken up in Section~\ref{sec:experiments}.

\section{Experiments}
\label{sec:experiments}

\begin{figure}[t]
\centering
\includegraphics[width=\textwidth]{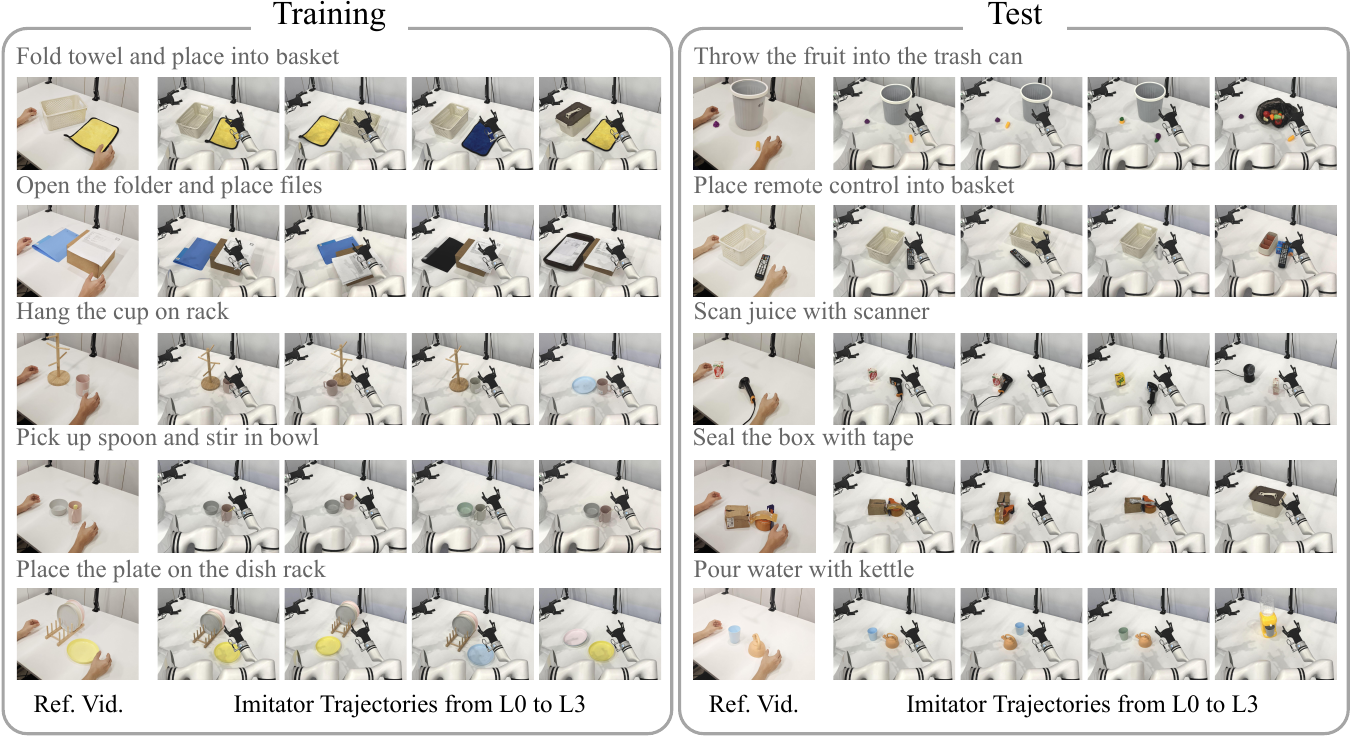}
\caption{\textbf{The experiment.} Evaluation spans training tasks (left) and held-out test tasks (right). 
For each task, the first image shows the human reference video, followed by representative imitator trajectories under increasing scene mismatch from L0 to L3. 
The real-world evaluation covers seen-task imitation, unseen zero-shot transfer, few-shot adaptation, and Arena human judgments.}
\label{fig:experiment}
\end{figure}

We evaluate human-video imitation along three axes: 
\textbf{(Q1)} the relative strength of VLAs, video-skill methods, and video-conditioned visuomotor agents,
\textbf{(Q2)} whether paired human--robot pretraining improves unseen-task transfer,
and
\textbf{(Q3)} where the L0--L3 hierarchy becomes difficult.
Across simulation, real-world rollouts, and Imitator Arena judgments, we find that video-based methods provide the strongest seen-task imitation, paired pretraining mainly improves few-shot adaptation rather than direct zero-shot transfer, and L3 affordance substitution is the clearest bottleneck.

\subsection{Setup}
\label{sec:setup}

\textbf{Tasks and protocol.}
We evaluate ten manipulation tasks across the L0--L3 hierarchy: five \emph{seen} tasks from the training set, and five disjoint \emph{unseen} tasks for zero-shot or few-shot transfer. The two evaluation groups are chosen to probe different things. The five seen tasks are the long-horizon or dexterous manipulation tasks -- stirring with a thin spoon, folding a towel, hanging a mug on a rack, pick up a paper, placing plates in a rack -- and ask how faithfully a model reproduces demanding \emph{learned} behavior given a human reference. The five unseen tasks are relatively simpler and closer to atomic skills (single-arm and bimanual placement, pouring, and an articulated folding task), but their demonstrations, objects, and layouts never appear training set, so they ask whether a new skill can be acquired from the reference alone. Both groups are run at all four levels, giving $40$ task--level pairs per domain.
Episodes are balanced across levels; example clips are shown in Figure \ref{fig:experiment}.
Task evaluation perturbs object placements, so success requires robust imitation rather than fixed-layout replay.
Simulation and the real-world robot scene are two \emph{separate} domains in IG-10K, and each model is trained and evaluated inside one domain only. But the sim and real domains share the same human-to-robot pairing, the same pretraining task split, and the same model configurations.

\textbf{IG-10K pretraining \& transfer.}
When using IG-10K pretraining, we pretrain on paired human-robot corpora of $15$, $30$, or $45$ tasks ($15 \subset 30 \subset 45$, and each budget was assembled to hold motion, object, and scene diversity as comparable as possible, so the corpus \emph{size} is the main thing to be considered in scaling), $50$ demonstrations per task are included in pretraining and excluding all unseen tasks. The task splits are given in Appendix~\ref{app:splits}. We evaluate seen tasks with models pre-trained on the three corpora. For unseen tasks we evaluate three settings: zero-shot transfer (\textbf{ZS}), in which pre-trained models directly execute on unseen tasks without any further parameter updates; from-scratch few-shot learning (\textbf{Scr.}), in which models are trained only on $10$ robot demonstrations of the five unseen tasks; and pre-train and fine-tune (\textbf{P+FT}), in which IG-10K-pre-trained models are fine-tuned on unseen tasks and $10$ demonstrations per task. Every level receives the same amount of supervision. 

\textbf{Baselines \& evaluation.}
We compare: (1) language-conditioned VLAs (GR00T-N1.6, RDT-1B, $\pi_{0.5}$, OpenVLA), (2) video-skill methods (XSkill, UniSkill), and (3) video-conditioned visuomotor models (ACT, Diffusion Policy, VQ-BeT) paired with DINOv2, SigLIP2, or VideoMAE encoders.
Simulation evaluates all policies and variants using success rate (SR) and sub-goal success rate (Sub-SR).
Four representative models ($\pi_{0.5}$, XSkill, ACT/DINOv2, and DP/DINOv2) on a dual-arm platform are evaluated in the real world, with $5$ trials per task-level pair against $10$ trials in simulation.
Imitator Arena further provides human success judgments, imitation score $Q$, and pairwise win rate (WR). Appendix Figure~\ref{fig:c7-validity} provides the alignment and validation of the two evaluation metrics.

\subsection{Q1: Which imitation interface is strongest?}
\label{sec:exp-video-language}

\begin{table}[t]
\centering
\small
\setlength{\tabcolsep}{2.7pt}
\caption{\textbf{Simulation results.} Seen, ZS, and P+FT are averaged over the $\{15,30,45\}$ task scales. 
\emph{Seen} = the five training tasks; \emph{ZS} = zero-shot evaluation on unseen tasks; \emph{Scr.} = trained from scratch on $10$ demonstrations of the unseen task, without IG-10K pre-training; \emph{P+FT} = IG-10K pre-trained and then fine-tuned on the same $10$ demonstrations; $\Delta = \text{P+FT} - \text{Scr.}$; \emph{Sub} = sub-goal success rate.}
\label{tab:main-sim}
\begin{tabular}{llcccccccccc}
\toprule
& & \multicolumn{2}{c}{Seen} & \multicolumn{2}{c}{ZS} & \multicolumn{2}{c}{Scr.} & \multicolumn{2}{c}{P+FT} & \multicolumn{2}{c}{$\Delta$} \\
\cmidrule(lr){3-4}\cmidrule(lr){5-6}\cmidrule(lr){7-8}\cmidrule(lr){9-10}\cmidrule(lr){11-12}
& Model & SR & Sub & SR & Sub & SR & Sub & SR & Sub & SR & Sub \\
\midrule
\multirow{4}{*}{\rotatebox{90}{VLA}}
& OpenVLA \citep{kim2024openvla} & 0.29 & 0.60 & 0.06 & 0.20 & 0.14 & 0.34 & 0.20 & 0.44 & +0.06 & +0.10 \\
& RDT-1B \citep{liu2025rdt} & 0.43 & 0.72 & 0.07 & 0.19 & 0.35 & 0.55 & 0.21 & 0.43 & --0.14 & --0.11 \\
& GR00T-N1.6 \citep{bjorck2025gr00t} & 0.51 & 0.81 & 0.03 & 0.17 & 0.21 & 0.49 & 0.67 & 0.79 & \cellcolor{yellow!50}\textbf{+0.46} & \cellcolor{yellow!50}\textbf{+0.31} \\
& $\pi_{0.5}$ \citep{intelligence2025pi_} & \cellcolor{yellow!50}\textbf{0.73} & \cellcolor{yellow!50}\textbf{0.89} & \cellcolor{yellow!50}\textbf{0.09} & \cellcolor{yellow!50}\textbf{0.22} & \cellcolor{yellow!50}\textbf{0.80} & \cellcolor{yellow!50}\textbf{0.86} & \cellcolor{yellow!50}\textbf{0.85} & \cellcolor{yellow!50}\textbf{0.91} & +0.05 & +0.04 \\
\midrule
\multirow{2}{*}{\rotatebox{90}{Skill}}
& UniSkill \citep{kim2025uniskill} & 0.75 & 0.89 & 0.07 & 0.17 & 0.27 & 0.46 & 0.59 & 0.76 & +0.32 & +0.30 \\
& XSkill \citep{xu2023xskill} & \cellcolor{yellow!50}\textbf{0.79} & \cellcolor{yellow!50}\textbf{0.91} & \cellcolor{yellow!50}\textbf{0.10} & \cellcolor{yellow!50}\textbf{0.25} & \cellcolor{yellow!50}\textbf{0.35} & \cellcolor{yellow!50}\textbf{0.51} & \cellcolor{yellow!50}\textbf{0.73} & \cellcolor{yellow!50}\textbf{0.82} & \cellcolor{yellow!50}\textbf{+0.38} & \cellcolor{yellow!50}\textbf{+0.31} \\
\midrule
\multirow{9}{*}{\rotatebox{90}{Video-VA}}
& DP/DINOv2 \citep{chi2025diffusion, oquab2023dinov2} & 0.67 & 0.84 & 0.07 & 0.21 & 0.13 & 0.36 & 0.54 & 0.73 & \cellcolor{yellow!50}\textbf{+0.41} & \cellcolor{yellow!50}\textbf{+0.37} \\
& DP/SigLIP2 \citep{tschannen2025siglip} & 0.61 & 0.84 & 0.05 & 0.20 & 0.11 & 0.35 & 0.48 & 0.69 & +0.38 & +0.34 \\
& DP/VideoMAE \citep{tong2022videomae} & 0.14 & 0.44 & 0.02 & 0.15 & 0.09 & 0.27 & 0.36 & 0.58 & +0.27 & +0.31 \\
& VQ-BeT/DINOv2 \citep{lee2024behavior} & 0.52 & 0.76 & \cellcolor{yellow!50}\textbf{0.13} & \cellcolor{yellow!50}\textbf{0.27} & 0.35 & 0.53 & 0.23 & 0.43 & --0.12 & --0.10 \\
& VQ-BeT/SigLIP2 & 0.57 & 0.78 & 0.08 & 0.23 & 0.26 & 0.46 & 0.17 & 0.34 & --0.08 & --0.11 \\
& VQ-BeT/VideoMAE & 0.30 & 0.51 & 0.07 & 0.21 & 0.14 & 0.36 & 0.15 & 0.33 & +0.01 & --0.04 \\
& ACT/DINOv2 \citep{zhao2023learning} & \cellcolor{yellow!50}\textbf{0.81} & \cellcolor{yellow!50}\textbf{0.93} & 0.02 & 0.13 & \cellcolor{yellow!50}\textbf{0.76} & \cellcolor{yellow!50}\textbf{0.83} & \cellcolor{yellow!50}\textbf{0.84} & \cellcolor{yellow!50}\textbf{0.88} & +0.09 & +0.05 \\
& ACT/SigLIP2 & 0.79 & 0.93 & 0.03 & 0.12 & 0.65 & 0.78 & 0.80 & 0.86 & +0.15 & +0.08 \\
& ACT/VideoMAE & 0.72 & 0.88 & 0.04 & 0.14 & 0.63 & 0.76 & 0.82 & 0.87 & +0.19 & +0.12 \\
\bottomrule
\end{tabular}
\end{table}

Table~\ref{tab:main-sim} and~\ref{tab:main-real} report the comprehensive performance in sim and real.
The results suggest that video-based pipelines are stronger for seen-task imitation but pre-training matters in few-shot learning.
In simulation, ACT/DINOv2 achieves the highest success rate (SR $=0.81$, Sub-SR $=0.93$), with XSkill and ACT/SigLIP2 close behind in seen-task evaluation; the strongest language-conditioned VLA, $\pi_{0.5}$, is lower (SR $=0.73$) in seen tasks but leads in few-shot adaptation (SR $=0.80$ in Scr. and SR $=0.85$ in P+FT). Appendix Figure~\ref{fig:c1-paradigm} gives an intuitive view of comprehensive simulation model performance.
In the real-world Arena, XSkill gives the strongest human-judged imitation among the representative models across all settings (SR $=0.63$ in Seen, SR $=0.29$ in ZS, SR $=0.28$ in Scr., SR $=0.49$ in P+FT), and video-based models constantly beat language-based models in Zero-shot transfer.
Within Video-VA, DINOv2 and SigLIP2 are consistently stronger than VideoMAE (Appendix Figure~\ref{fig:c2-encoder}), indicating that the video representation is also an important design choice.

\subsection{Q2: Does scale improve zero-shot transfer or few-shot adaptation?}
\label{sec:exp-scaling}

In simulation, paired pretraining mainly supports few-shot adaptation. Zero-shot performance remains near the floor across paradigms, with the best simulation result reaching only $\mathrm{SR}=0.13$. Current frameworks still struggle to generalize to unseen tasks without task-specific robot data. In contrast, P+FT improves few-shot learning for most models, with benefits growing with scale: from $15$ to $45$ pretraining tasks, P+FT success improves for $14$ of $15$ simulation variants (Appendix Figure~\ref{fig:c3-scaling}), and P+FT wins over Scr. for $12$ of the $15$ simulation variants (Table~\ref{tab:main-sim}). In the real-world evaluation, all four representative models improve in the same way in few-shot adaptation, and the three video-conditioned ones also gain some zero-shot ability as the corpus grows, while the language-conditioned $\pi_{0.5}$ does not (Appendix Figure~\ref{fig:c3-scaling}). According to Appendix~\ref{app:q2-levelscale}, Table~\ref{tab:level-scale}, P+FT success rises from $15$ to $45$ pre-training tasks at every one of L0--L3, in both domains. Robust zero-shot imitation from human video remains difficult, but more paired data does make a model easier to adapt.

\subsection{Q3: Where does the hierarchy become hard?}
\label{sec:exp-levels}

We analyze the hierarchy in the real-world P+FT setting, where policies receive task-specific adaptation and are judged by humans.
Table~\ref{tab:perlevel} shows the main difficulty appears at L3.
Averaged over the four representative models, SR is nearly stable from L0 to L2 (around 0.4) but drops to 0.29 at L3; the human imitation score follows suit, falling from about 6.2 on L0--L2 to 5.62 on L3. This drop is clearest for the strongest L0--L2 policy: XSkill holds SR $\approx 0.53$--$0.57$ through L2 but falls to 0.29 at L3, while other policies show a flatter profile from a lower operating point. Current policies therefore cope with a rearranged layout, and with a replacement object of the same kind, but not with an object that has to be used in a different way to reach the same end.

\begin{table}[t]
\centering
\scriptsize
\setlength{\tabcolsep}{2.2pt}
\caption{\textbf{Real-world Arena results} combining seen performance and unseen transfer. Seen, ZS, and P+FT are averaged over the $\{15,30,45\}$ task scales. Settings as in Table~\ref{tab:main-sim}. $\overline{Q}$ = mean human imitation score on the $0$--$10$ rubric of Appendix~\ref{app:arena}; WR = win rate in blind A/B comparisons. On hardware there is no automated metric, so SR here is the human success judgment.}
\label{tab:main-real}
\resizebox{\textwidth}{!}{%
\begin{tabular}{llccccccccccccccc}
\toprule
& & \multicolumn{3}{c}{Seen} & \multicolumn{3}{c}{ZS} & \multicolumn{3}{c}{Scr.} & \multicolumn{3}{c}{P+FT} & \multicolumn{3}{c}{$\Delta$} \\
\cmidrule(lr){3-5}\cmidrule(lr){6-8}\cmidrule(lr){9-11}\cmidrule(lr){12-14}\cmidrule(lr){15-17}
& Model & SR & $\overline{Q}$ $\uparrow$ & WR $\uparrow$ & SR & $\overline{Q}$ $\uparrow$ & WR $\uparrow$ & SR & $\overline{Q}$ $\uparrow$ & WR $\uparrow$ & SR & $\overline{Q}$ $\uparrow$ & WR $\uparrow$ & SR & $\overline{Q}$ $\uparrow$ & WR $\uparrow$ \\
\midrule
& $\pi_{0.5}$ \citep{intelligence2025pi_} & \underline{0.51} & \underline{6.87} & \underline{0.56} & 0.04 & 2.34 & 0.04 & \underline{0.27} & \textbf{5.52} & \underline{0.50} & \underline{0.36} & \underline{5.89} & \underline{0.73} & +0.09 & +0.37 & +0.23 \\
& XSkill \citep{xu2023xskill} & \textbf{0.63} & \textbf{7.49} & \textbf{0.89} & \textbf{0.29} & \underline{5.24} & \textbf{0.50} & \textbf{0.28} & \underline{5.40} & \textbf{0.52} & \textbf{0.49} & \textbf{6.72} & \textbf{0.95} & \textbf{+0.21} & \textbf{+1.32} & \underline{+0.43} \\
& ACT/DINOv2 \citep{zhao2023learning, oquab2023dinov2} & 0.49 & 6.70 & 0.41 & \underline{0.26} & \textbf{5.44} & \underline{0.42} & 0.22 & 5.38 & 0.24 & 0.35 & 5.88 & 0.70 & \underline{+0.13} & \underline{+0.50} & \textbf{+0.46} \\
& DP/DINOv2 \citep{chi2025diffusion} & 0.39 & 6.19 & 0.14 & 0.22 & 4.96 & 0.25 & 0.22 & 5.32 & 0.24 & 0.31 & 5.77 & 0.58 & +0.09 & +0.45 & +0.34 \\
\bottomrule
\end{tabular}%
}
\end{table}

\begin{wraptable}{r}{0.62\textwidth}
\centering
\vspace{-\baselineskip}
\scriptsize
\setlength{\tabcolsep}{4pt}
\caption{\textbf{Real-world P+FT success by hierarchy level.} $\mathrm{SR}$ and human imitation score $\overline{Q}$, computed from real-world pretrain+finetune trials and averaged over the $\{15,30,45\}$-task scales. The corresponding line plot is Appendix Figure~\ref{fig:c5-perlevel}.}
\label{tab:perlevel}
\begin{tabular}{lcccccccc}
\toprule
& \multicolumn{2}{c}{L0} & \multicolumn{2}{c}{L1} & \multicolumn{2}{c}{L2} & \multicolumn{2}{c}{L3} \\
\cmidrule(lr){2-3}\cmidrule(lr){4-5}\cmidrule(lr){6-7}\cmidrule(lr){8-9}
Model & SR & $\overline{Q}$ $\uparrow$ & SR & $\overline{Q}$ $\uparrow$ & SR & $\overline{Q}$ $\uparrow$ & SR & $\overline{Q}$ $\uparrow$ \\
\midrule
$\pi_{0.5}$ \citep{intelligence2025pi_} & 0.44 & 6.27 & 0.35 & 5.77 & 0.37 & 5.97 & 0.28 & 5.57 \\
XSkill \citep{xu2023xskill} & 0.53 & 6.80 & 0.57 & 7.20 & 0.56 & 7.23 & 0.29 & 5.63 \\
ACT/DINOv2 \citep{zhao2023learning, oquab2023dinov2} & 0.37 & 6.00 & 0.37 & 5.90 & 0.33 & 5.77 & 0.33 & 5.83 \\
DP/DINOv2 \citep{chi2025diffusion} & 0.33 & 5.97 & 0.37 & 6.00 & 0.29 & 5.70 & 0.25 & 5.43 \\
\midrule
Average & 0.42 & 6.26 & 0.42 & 6.22 & 0.39 & 6.17 & 0.29 & 5.62 \\
\bottomrule
\end{tabular}
\end{wraptable}

\section{Conclusion}
\label{sec:discussion}

The Imitator Game Benchmark studies robot video imitation under progressively increasing mismatch between the human demonstration and the robot's own scene. To support this setting, we introduce IG-10K, a large-scale environment-aligned human--robot paired dataset spanning four levels of imitation, together with Imitator Arena for structured human evaluation of imitation behavior. Our experiments across multiple policy families and training regimes show that current systems transfer poorly to unseen tasks without task-specific robot data, and struggle when imitation requires functional substitution and intent-level adaptation. These suggest that intent-level imitation is unlikely to arrive as a by-product of scaling either the corpus or the action head only, and that the productive direction is an interface that keeps the demonstration available as evidence about \emph{purpose} -- affordance-aware object grounding, explicit goal inference, or intermediate representations that survive object substitution -- rather than as a specification to be transcribed once and executed. IG-10K, Imitator Arena, and the L0--L3 protocol are released so that such interfaces can be measured on the rung where they would actually differ.

\paragraph{Limitations.}
IG-10K remains finite and manually designed, leaving broader scaling and more open-ended transfer substitutions to future work. Each task--level pair currently carries a single substitution pattern, which buys comparability across levels at the cost of not measuring variance over substitutions, and the hierarchy grades scene mismatch while holding the demonstrator--robot embodiment gap approximately fixed. Finite assets and rule-based planned motion further bias the simulated L3 substitution results. The evaluated policies are adapted to the Imitator Game rather than designed for intent-level human-video imitation, the best suitable framework for the Imitator Game remains an open question. Current models will fail in cases mainly in unseen zero-shot generalization tasks and functional substitution, we will discuss it in Appendix \ref{app:case-study}.

\newpage
\bibliography{main}

\clearpage
\appendix

\section{The Imitator Capability: Supporting Detail}
\label{app:capability}

\subsection{Scene-Decomposition Details}
\label{app:scene-decomposition}

This appendix expands the scene decomposition of Section~\ref{sec:problem-setting}. Recall that a demonstration is recorded in $\mathcal{S}_{\text{demo}}$ and the robot acts in $\mathcal{S}_{\text{imit}}$. We write a scene as $\mathcal{S}=(\{O_i\},P)$, where $P$ denotes the spatial configuration, and each task-relevant object $O_i=(A_i,G_i,S_i)$ has appearance $A_i$, geometry $G_i$, and semantic category $S_i$. The appendix clarifies when $P$ is comparable across the two scenes and how the benchmark uses the construction rule $S \!\Rightarrow\! G \!\Rightarrow\! A$ to define the level hierarchy.

\paragraph{When is $P$ a meaningful comparison dimension?}
The spatial configuration $P$ is only a meaningful comparison dimension between $\mathcal{S}_{\text{demo}}$ and $\mathcal{S}_{\text{imit}}$ when the two scenes share the same object set. Once the objects differ, there is nothing to line up one-to-one: a teapot's pose and a water tap's pose are not two values of the same quantity, so $P$ is not ``different'' -- the question simply does not arise. This is why, in the level table of Section~\ref{sec:levels}, the $P$ column is marked ``n/a'' and not ``$\times$'' at the levels where the object set has changed. For the same reason there is no level that changes $P$ \emph{and} the object set at once: such a level could not be described in these terms at all.

\paragraph{Edge cases of the implication chain.}
The assumption $S \!\Rightarrow\! G \!\Rightarrow\! A$ has rare counterexamples in which semantic category and physical form come apart -- a wax banana (the semantics of ``banana'' without its edibility or, depending on construction, its exact geometry), or a decorative object shaped like a cup that is in fact a candle holder (the geometry of a cup with the semantics of a candle holder). In such cases the conventional link between category and form is deliberately broken for ornamental or deceptive purposes. We treat these as out of scope for benchmark instances: every task-relevant object in IG-10K is a functional exemplar of its semantic category, so that the chain holds, and object-substitution levels can be defined by a single intrinsic change. We list these cases to state the assumption precisely; none of them occurs in IG-10K at this stage.

\subsection{The Four Levels: Per-Level Detail and Psychological Grounding}
\label{app:levels}

This appendix expands Section~\ref{sec:levels} with the full per-level scene relations and worked examples, the L2 appearance-only sub-case, the correspondence to developmental-psychology constructs, and the relationship of our scene-mismatch hierarchy to embodiment-gap hierarchies.

\paragraph{L0 -- Trajectory imitation.} \emph{Scene relation.} The object set is intrinsically identical ($A$ and $G$ preserved on every task-relevant object) and $P$ is preserved in the operational sense of Section~\ref{sec:levels}--the trajectory implicit in $V$ would still succeed in $\mathcal{S}_{\text{imit}}$. \emph{What the scene permits.} The demonstrated end-effector motion and object motion remain feasible. \emph{What the imitator is graded on.} The rollout should preserve the demonstrated motion pattern as closely as the shared scene permits, while the benchmark still scores the episode through the Arena judgments and task-completion metrics rather than a hand-designed trajectory-distance metric. \emph{Example.} Same teapot and cup in the same positions; the imitator pours along essentially the demonstrated motion.

\paragraph{L1 -- Object end-state imitation.} \emph{Scene relation.} The object set is intrinsically identical but $P$ has changed -- a literal replay of the trajectory implicit in $V$ would no longer succeed. \emph{What the scene permits.} The demonstrated trajectory is no longer correct, but exactly the same objects can be brought to exactly the same final states. \emph{What the imitator is graded on.} Match of final object states with $V$; trajectory similarity is not rewarded. \emph{Example.} Same teapot and cup, but the cup has moved to the opposite side of the table; the imitator should still leave the cup full of water.

\paragraph{L2 -- Semantic-task imitation.} \emph{Scene relation.} At least one task-relevant object has changed in $A$ (and possibly $G$); each substitution preserves the object's semantic category $C$. \emph{What the scene permits.} Because the objects differ, the precise final states of $V$ are not reproducible (a handled mug is not a handleless cup), but the semantic task -- ``make a cup of tea,'' ``set the table,'' ``stir the soup'' -- remains well-defined. \emph{What the imitator is graded on.} Completion of the semantic task implied by $V$, judged at an abstraction that survives object substitution. \emph{Example.} The demonstrator pours from a ceramic teapot into a handleless cup; the imitator's scene has a glass teapot and a handled mug, and the expected behavior is still ``pour tea from the teapot into the mug.''

\paragraph{The L2 appearance-only sub-case.} Within L2 it is sometimes useful to distinguish the sub-case in which only $A$ changes while $G$ is preserved (a ceramic cup replaced by a glass cup of identical shape) from the sub-case in which both $A$ and $G$ change (the cup replaced by a mug of different shape). The first isolates a pure perceptual invariance -- the policy must recognize that two visually different objects play the same functional role -- while the second additionally demands that the policy adapt its grasp and motion to a different shape. We report both sub-cases where a task admits them, but treat them as a single L2 level for headline metrics because both share the same fidelity ceiling: the semantic task rather than per-object final states.

\paragraph{L3 -- Affordance-adapted imitation.} \emph{Scene relation.} At least one task-relevant object changes in semantic category $C$; under our benchmark construction this also changes geometry $G$ and appearance $A$. \emph{What the scene permits.} The object-specific execution shown in $V$ is no longer directly reproducible, but the underlying purpose can still be served by selecting an object in $\mathcal{S}_{\text{imit}}$ whose affordance can be adapted to the same end. \emph{What the imitator is graded on.} Achievement of an action that satisfies the underlying intent of $V$ via affordance adaptation, assessed by human evaluators in the Arena (Section~\ref{sec:arena}); neither trajectory nor specific final states are expected to resemble $V$. \emph{Examples.} (i) The demonstrator drives a nail with a hammer; the imitator's scene contains no hammer but a stone, and the expected behavior is to use the stone as a hammer. (ii) The demonstrator steeps a tea bag to make a hot drink; the imitator's scene contains no tea bag but an instant drink powder, so the expected behavior is to make a drink by mixing rather than steeping.

\paragraph{Correspondence to developmental-psychology constructs.} The four fidelity levels take operational inspiration from a hierarchy of social-learning constructs: surface-motor copying without attribution of instrumental efficacy (``mimicking'' in Tomasello's sense~\citep{whiten2004apes,tomasello2009cultural}), end-state emulation~\citep{wood1996social,tomasello2009cultural,whiten2009emulation}, goal-directed (intentional) imitation~\citep{meltzoff1995understanding,tomasello2009cultural,gergely2002rational}, and analogical transfer of an observed practice to functionally equivalent means~\citep{brown1989analogical,chen2000across,byrne1998learning}. We adopt these as labels for our hierarchy of replication fidelities rather than as claims about the cognitive mechanisms operating in our system, and in particular do not claim that our system attributes mental states. The four levels are our synthesis: to our knowledge, no single published taxonomy enumerates exactly these four, and the framing builds on Tomasello's mimicking/emulation/imitation distinction and Byrne and Russon's~\citep{byrne1998learning} program-level imitation. What we actually measure is simpler than any of these constructs: as the robot's scene moves further from the demonstration, what an imitator can still reproduce becomes less and less specific.

\paragraph{Relationship to embodiment-gap hierarchies.} Our hierarchy is orthogonal to embodiment-gap hierarchies such as RHyME's~\citep{kedia2025one}, which grade the mismatch between the demonstrator's body and the robot's body while holding the scene fixed. The Imitator Game holds the demonstrator--robot embodiment gap approximately fixed and instead grades the scene-level mismatch between $\mathcal{S}_{\text{demo}}$ and $\mathcal{S}_{\text{imit}}$. We likewise assume a visual front-end invariant to lighting and viewpoint variation between $V$ and the imitator's observations, treating these as out of scope rather than as additional levels. The two axes -- embodiment gap and scene-level mismatch -- are complementary, and combining them is a natural direction for future benchmarks.

\section{The Imitator Benchmark: Supporting Detail}
\label{app:benchmark}

\subsection{IG-10K Collection and Annotation Details}
\label{app:dataset}

\begin{figure}[t]
\centering
\includegraphics[width=\textwidth]{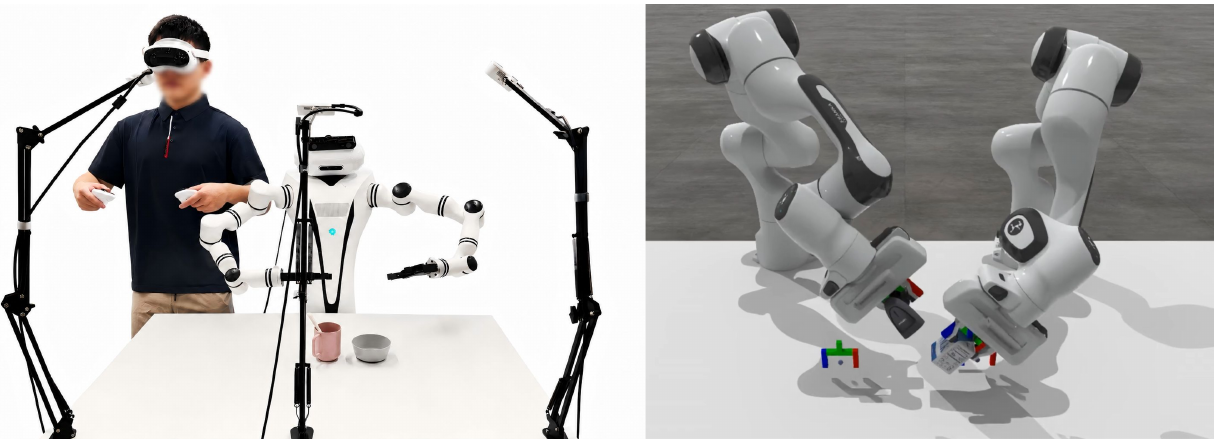}
\caption{\textbf{IG-10K data collection.} \textit{(left)} The real-world data are collected with VR teleoperation on a dual-arm robot platform, with visual data recorded by three third-view RealSense cameras and one front-facing ZED2i RGB camera. \textit{(right)}  simulation data are collected with handcrafted motion-planning waypoints and trajectories planned using OMPL.}
\label{fig:platform}
\end{figure}

This appendix expands Section~\ref{sec:dataset} with the full task suite, the data-collection platforms and acquisition parameters, the sensing rig, and the multi-modal annotation toolchain. Figure~\ref{fig:dataset} shows representative paired episodes, the object assets, and the offline annotation layers; Figure~\ref{fig:platform} shows the collection rig.

\paragraph{Data collection.} Real-robot data are collected on a dual-arm Realman platform, teleoperated via VR, with full joint angles and 6-DoF end-effector poses recorded at a control frequency of 30 Hz. Simulation data are collected on a dual-arm Franka in ManiSkill3, with assets drawn from YCB, PartNet-Mobility, RoboTwin, and Sketchfab; trajectories are planned by OMPL from hand-crafted waypoints. Of the $20{,}000+$ paired episodes in total, 11.7K are real human-robot paired episodes and 10K are paired simulation episodes. Each real paired episode contains one human clip and one corresponding real-robot clip; the simulation episodes provide paired interactive environments with fully observable automated metrics. Per task--level pair we collect 50 paired episodes on average.

\paragraph{Sensing rig.} All human, real-world robot, and simulation scenes are captured by a wide-angle ZED2i ego-centric camera together with three RealSense third-person cameras (left-frontal, top-down, right-frontal), all providing RGB-depth vision. For model input we standardize on the ZED2i RGB stream for compatibility with the diverse pre-training setups of the baseline families (Section~\ref{sec:baselines}); the remaining views and the depth channels are retained in the released data for use by future methods.

\paragraph{Task suite.} Table~\ref{tab:tasksuite} lists all 50+ base tasks by domain, together with the levels at which each is instantiated. A task omits L3 only where affordance substitution would be semantically ill-defined or physically unsafe; the omitted cases and their justification are noted in the table.

\begin{table}[!t]
\centering
\small
\setlength{\tabcolsep}{5pt}
\caption{\textbf{The IG-10K task suite.}
  All base tasks by domain, with the levels at which each is instantiated
  (\checkmark\ present, -- omitted).
  Per-domain counts: Household~12, Supermarket~7, Restaurant~17,
  Logistics~4, Hospital~6, Laboratory~7.}
\label{tab:tasksuite}
\begin{tabular}{llccccl}
\toprule
Domain & Task & L0 & L1 & L2 & L3 & Notes \\
\midrule
\multirow{12}{*}{Household (12)}
 & \textsc{H1: Stir Soup in Bowl}           & \checkmark & \checkmark & \checkmark & \checkmark & \\
 & \textsc{H2: Return Book to Shelf}         & \checkmark & \checkmark & \checkmark & \checkmark & \\
 & \textsc{H3: Fold Clothes into Basket}     & \checkmark & \checkmark & \checkmark & \checkmark & \\
 & \textsc{H4: Return Remote to Box}         & \checkmark & \checkmark & \checkmark & \checkmark & \\
 & \textsc{H5: Put Magazine in Folder}       & \checkmark & \checkmark & \checkmark & \checkmark & \\
 & \textsc{H6: Return Body Wash to Holder}   & \checkmark & \checkmark & \checkmark & \checkmark & \\
 & \textsc{H28: Put Shoes in Shoe Box}       & \checkmark & \checkmark & \checkmark & \checkmark & \\
 & \textsc{H35: Place Brush on Stand}        & \checkmark & \checkmark & \checkmark & \checkmark & \\
 & \textsc{H49: Fold Towel into Basket}      & \checkmark & \checkmark & \checkmark & \checkmark & \\
 & \textsc{H50: Juice with Juicer}           & \checkmark & \checkmark & \checkmark & \checkmark & \\
 & \textsc{H57: Hang Cup on Rack}            & \checkmark & \checkmark & \checkmark & \checkmark & \\
 & \textsc{H58: Fix Wood with Screwdriver}   & \checkmark & \checkmark & \checkmark & \checkmark & \\
\midrule
\multirow{7}{*}{Supermarket (7)}
 & \textsc{H7: Put Apple in Basket}          & \checkmark & \checkmark & \checkmark & \checkmark & \\
 & \textsc{H8: Sort Fruits into Baskets}     & \checkmark & \checkmark & \checkmark & \checkmark & \\
 & \textsc{H9: Weigh Apple on Scale}         & \checkmark & \checkmark & \checkmark & \checkmark & \\
 & \textsc{H10: Stock Chips on Shelf}        & \checkmark & \checkmark & \checkmark & \checkmark & \\
 & \textsc{H11: Organize Items on Shelf}     & \checkmark & \checkmark & \checkmark & \checkmark & \\
 & \textsc{H12: Pack Fruits in Shopping Bag} & \checkmark & \checkmark & \checkmark & \checkmark & \\
 & \textsc{H13: Scan Beverage Barcode}       & \checkmark & \checkmark & \checkmark & \checkmark & \\
\midrule
\multirow{17}{*}{Restaurant (17)}
 & \textsc{H14: Place Plates in Rack}        & \checkmark & \checkmark & \checkmark & \checkmark & \\
 & \textsc{H15: Chop Carrot}                 & \checkmark & \checkmark & \checkmark & \checkmark & \\
 & \textsc{H16: Arrange Fruits on Plate}     & \checkmark & \checkmark & \checkmark & \checkmark & \\
 & \textsc{H17: Add Sauce to Burger}         & \checkmark & \checkmark & \checkmark & \checkmark & \\
 & \textsc{H18: Wash Cookware}               & \checkmark & \checkmark & \checkmark & \checkmark & \\
 & \textsc{H19: Wipe Table After Meal}       & \checkmark & \checkmark & \checkmark & \checkmark & \\
 & \textsc{H20: Pour Water into Cup}         & \checkmark & \checkmark & \checkmark & \checkmark & \\
 & \textsc{H21: Discard Food Waste}          & \checkmark & \checkmark & \checkmark & \checkmark & \\
 & \textsc{H22: Transfer Food with Spoon}    & \checkmark & \checkmark & \checkmark & \checkmark & \\
 & \textsc{H23: Weigh Ingredients}           & \checkmark & \checkmark & \checkmark & \checkmark & \\
 & \textsc{H24: Staple Documents}            & \checkmark & \checkmark & \checkmark & \checkmark & \\
 & \textsc{H25: Stir Ingredients in Bowl}    & \checkmark & \checkmark & \checkmark & \checkmark & \\
 & \textsc{H45: Filter Liquid}               & \checkmark & \checkmark & \checkmark & \checkmark & \\
 & \textsc{H46: Pack Food in Box}            & \checkmark & \checkmark & \checkmark & \checkmark & \\
 & \textsc{H47: Set Tableware}               & \checkmark & \checkmark & \checkmark & \checkmark & \\
 & \textsc{H48: Open Lid to Inspect Pot}     & \checkmark & \checkmark & \checkmark & \checkmark & \\
 & \textsc{H56: Place Burger in Meal Tray}   & \checkmark & \checkmark & \checkmark & \checkmark & \\
\midrule
\multirow{4}{*}{Logistics (4)}
 & \textsc{H26: Sort and Seal Goods}         & \checkmark & \checkmark & \checkmark & \checkmark & \\
 & \textsc{H27: Identify Barcode and Place}  & \checkmark & \checkmark & \checkmark & \checkmark & \\
 & \textsc{H29: Tear Packaging for Sample}   & \checkmark & \checkmark & \checkmark & --         & ill-defined \\
 & \textsc{H30: Seal and Pack Box}           & \checkmark & \checkmark & \checkmark & \checkmark & \\
\midrule
\multirow{6}{*}{Hospital (6)}
 & \textsc{H31: Sort Medications}            & \checkmark & \checkmark & \checkmark & \checkmark & \\
 & \textsc{H32: File Medical Documents}      & \checkmark & \checkmark & \checkmark & \checkmark & \\
 & \textsc{H33: Pour Medication Liquid}      & \checkmark & \checkmark & \checkmark & --         & hazardous \\
 & \textsc{H34: Put Pills in Pill Box}       & \checkmark & \checkmark & \checkmark & \checkmark & \\
 & \textsc{H36: Place Instruments in Tray}   & \checkmark & \checkmark & \checkmark & --         & ill-defined \\
 & \textsc{H37: Clean Container Opening}     & \checkmark & \checkmark & \checkmark & \checkmark & \\
\midrule
\multirow{7}{*}{Laboratory (7)}
 & \textsc{H38: Stir Liquid in Beaker}       & \checkmark & \checkmark & \checkmark & --         & hazardous \\
 & \textsc{H39: Pour Liquid Reagent}         & \checkmark & \checkmark & \checkmark & --         & hazardous \\
 & \textsc{H40: Weigh Solid Material}        & \checkmark & \checkmark & \checkmark & \checkmark & \\
 & \textsc{H41: Filter Solution}             & \checkmark & \checkmark & \checkmark & \checkmark & \\
 & \textsc{H42: Wash Laboratory Vessels}     & \checkmark & \checkmark & \checkmark & \checkmark & \\
 & \textsc{H43: Open Container Lid}          & \checkmark & \checkmark & \checkmark & --         & hazardous \\
 & \textsc{H44: Grind Solid Substance}       & \checkmark & \checkmark & \checkmark & \checkmark & \\
\bottomrule
\end{tabular}
\end{table}

\paragraph{Multi-modal annotation.} Each video carries three annotation layers. (i) \emph{Language}: structured descriptions of the demonstrated task at three levels of abstraction--per-arm sub-task sequence, semantic task description, and underlying intent--authored by human annotators and augmented with Qwen3-VL. These three abstraction levels are precisely what the VLA captioner of Section~\ref{sec:baselines} consumes. (ii) \emph{Hand pose}: 3D MANO hand pose estimated by WiLoR (per-view; see Appendix~\ref{app:mano-processing}). (iii) \emph{Segmentation}: semantic segmentation masks from Grounded-SAM-2, conditioned on per-video object inventories. The mask layer is an offline annotation produced for analysis and future training objectives rather than a component of the scoring pipeline; its generation procedure and quality control are described separately in Appendix~\ref{app:mask-generation}.

\paragraph{Constructing the L2 sub-cases.} The appearance-only and shape-changing sub-cases of L2 (Appendix~\ref{app:levels}) are built by controlled substitution. For the appearance-only sub-case we replace a task-relevant object with one of the same semantic category and the same geometry but different appearance (e.g.\ a ceramic cup and a glass cup sharing a mesh up to material), isolating perceptual invariance. For the shape-changing sub-case we additionally vary geometry within the category (e.g.\ a handleless cup and a handled mug), so the policy must also adapt grasp and motion.

\paragraph{L3-omitted tasks.} A small number of tasks omit L3 where affordance substitution would be semantically ill-defined or unsafe; these are marked in Table~\ref{tab:tasksuite}, the laboratory hazardous-liquid cases being the canonical example.

\subsection{Corpus Construction, Splits, and the Simulation--Real Relation}
\label{app:splits}

This appendix expands the split description of Section~\ref{sec:dataset} and states exactly which data are used to pre-train the models evaluated in Section~\ref{sec:experiments}.

\paragraph{Two domains, one experimental structure.} Simulation and the real world are separate domains. The embodiments differ (dual-arm Franka in ManiSkill3 versus dual-arm Realman), and the visual sensor statistics differ, so we never mix the two corpora: simulation models are trained on simulation data and evaluated in simulation, real-world models are trained on real data and evaluated on hardware. Everything else is held common. Both domains use the same model configurations and hyper-parameters, the same input/output interface (the ZED2i RGB stream as input; dual-arm $7$-DoF joint commands plus gripper as output), and the same \texttt{lerobot} data format, so a single training codebase produces both sets of variants. Both use the same human-to-robot pairing: one human demonstration video is paired with the robot episodes of the corresponding task, and the same human clip serves all four levels of that task. Both use the same task split, listed below. A small number of real tasks have no simulated counterpart, because the required objects or contact behaviors are not reproducible with our simulation assets; these are the only points at which the two domains' corpora differ, and the $15/30/45$ nesting and the seen/unseen assignment are preserved in each domain. Simulated paired episodes do not use a separately recorded human demonstration. A simulated episode reuses the \emph{real} human clip recorded for the same task, with only the imitator scene replaced by its simulated counterpart. MANO hand pose is provided only for real human clips.

\begin{table}[th]
\centering
\small
\setlength{\tabcolsep}{4pt}
\caption{\textbf{Pre-training corpora and the unseen split.} The three corpora are nested; the $30$-task corpus adds the listed tasks to the $15$-task corpus, and the $45$-task corpus adds its own to the $30$-task corpus. \textbf{Bold} marks the five seen evaluation tasks. The five unseen tasks appear in no corpus.}
\label{tab:corpora}
\begin{tabular}{@{}l>{\raggedright\arraybackslash}p{0.85\textwidth}@{}}
\toprule
Corpus & Tasks \\
\midrule
$\mathcal{C}_{15}$ & \textbf{StirSpoon}, PlaceClothBasket, PlaceMagazineFolder, PickWash, PlaceChipsRack, PlaceFruitBox, \textbf{PlacePlateRack}, CutFruit, \textbf{PlaceFileFolder}, PlaceBrushRest, CleanCup, GrindFood, LiftLidFromSkillet, \textbf{FoldTowel}, \textbf{PlaceMugRack} \\
\addlinespace[2pt]
$\mathcal{C}_{30}$ (adds) & PlaceCommodityRack, PourKetchupFries, WipePot, CleanDesk, TransFood, PickTennisBallGolfBall, PickPillToRegions, PourLiquidCup, PlaceCupPlate, PutCubeOnScale, PourCup, PutBox, KnifeBowlFork, PressJuicer, PlaceScrewdriver \\
\addlinespace[2pt]
$\mathcal{C}_{45}$ (adds) & PlaceBookBookcase, PickAppleBasket, PickAppleBananaToBaskets, PickAppleToScale, PickFruitsToPlate, PlaceFoodScale, PressStapler, ScanPillBottle, PlaceShoeBox, OpenBox, PlacePillBox, PourLiquidMug, OpenLiquidCap, PourLiquidFilter, PlaceBurgerTray \\
\midrule
Unseen & PickRemoteControl, ScanMilkBox, PourKettle, PickFood, FoldBox \\
\bottomrule
\end{tabular}
\end{table}

\paragraph{Nested pre-training corpora.} The three pre-training corpora are nested, $\mathcal{C}_{15} \subset \mathcal{C}_{30} \subset \mathcal{C}_{45}$, and were assembled so that motion types, manipulated objects, scene domains, and horizon lengths stay as diverse as each budget permits. The intent is that corpus \emph{scale} is the variable that changes between them, rather than diversity or task character; scaling results (Section~\ref{sec:exp-scaling}) should be read against this construction. The five unseen tasks are excluded from all three corpora. Table~\ref{tab:corpora} lists the membership.

\paragraph{Seen and unseen evaluation tasks.} Evaluation uses five seen tasks drawn from $\mathcal{C}_{15}$ and five unseen tasks. The seen tasks are the long-horizon or dexterous members of the corpus -- stirring with a thin spoon, folding a towel, hanging a mug on a rack, filing a folder, and placing plates in a rack -- and test how faithfully a model reproduces demanding learned behavior given a human reference. The unseen tasks are individually simpler and closer to atomic skills (single-arm placement, bimanual scanning, bimanual placement, pouring, and an articulated folding task), but their demonstrations, objects, and layouts are absent from every corpus, so they test acquisition of a new skill from the reference alone rather than recall. Both groups are evaluated at all four levels, giving $40$ task--level pairs per domain, with $10$ trials per pair in simulation and $5$ on hardware.

\paragraph{Substitution patterns and supervision per task--level pair.} Each task--level pair uses exactly one substitution pattern. The number of possible substitutions grows as the level rises, so allowing more of them at L3 than at L0 would mean comparing how varied the substitutions are instead of how well the policies handle them. Each pair receives $50$ pre-training demonstrations, collected with small position variations for robustness. Supervision is therefore equal across levels, and the $10$ few-shot demonstrations are not a budget divided among the four levels. Broadening the substitution patterns per level is left to future versions of IG-10K.

\subsection{Imitator Arena: Scoring Procedure and Metric Definitions}
\label{app:arena}

\begin{figure}[t]
\centering
\includegraphics[width=\textwidth]{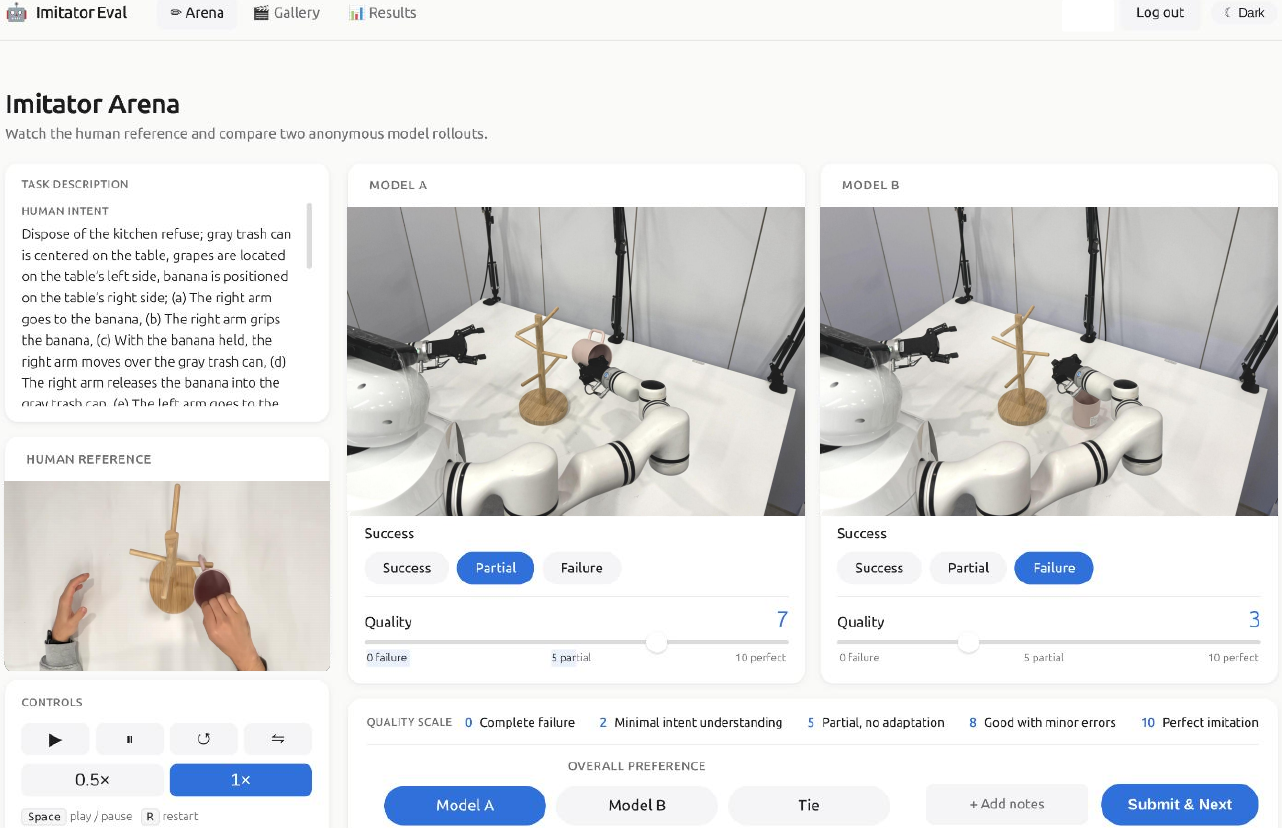}
\caption{\textbf{The Imitator Arena.} The web interface shows the task description and human reference video on the left, and two anonymized model rollouts from the same imitator scene. Human evaluators judge each rollout for task completion (SUCCESS/PARTIAL/FAILURE), assign an imitation-quality score on a fixed 0–10 scale, and make an overall pairwise preference or tie decision. Level labels and model identities are hidden to focus evaluation on imitation of the demonstrated intent.}
\label{fig:arena-platform}
\end{figure}

This appendix expands Section~\ref{sec:arena} with the platform and annotation protocol, the imitation-quality rubric, the pairwise-comparison protocol, the full execution-and-scoring procedure, the formal metric definitions with a worked per-task example, and annotation reliability.

\paragraph{Platform and annotation protocol.} Imitator Arena is a web platform on which each annotation screen presents three time-synchronized videos--the demonstration $V$ and two anonymized, randomly ordered rollouts on the same imitator scene--and collects the three judgments of Section~\ref{sec:arena}. Annotation is carried out by a pool of $10$ human evaluators who divide the workload roughly evenly, so that each annotation screen is judged by one human evaluator and each human evaluator contributes approximately one tenth of the total. The evaluators are volunteers with no involvement in the project; no crowdsourcing platform was used, participation was voluntary and uncompensated by task, and no personal data were collected, so the study raises no additional ethical concerns. Restricting a screen to a single evaluator is what keeps judgments independent and lets the platform double as an anonymous public leaderboard, and reliability is obtained by repetition rather than by co-annotation of a screen: screens are drawn at random from the eligible pool, so at the study's volume each demonstration--rollout pair is judged $3$--$5$ times, by different evaluators, on independently sampled screens. Supporting multiple annotators per screen, and reporting an explicit inter-annotator agreement statistic alongside the qualification pass, is a planned extension of the platform. In total the human study comprises $15{,}000$ pairwise comparisons in simulation and $5{,}000$ on real hardware, drawn from $36{,}000$ distinct simulation rollouts and $4{,}800$ distinct real-world rollouts (the per-bucket breakdown is given below). Model identities and imitation levels are never shown, and the left/right order of the two rollouts is randomized to control position bias.

\paragraph{Quality rubric.} The per-rollout imitation-quality score $q \in [0,10]$ (question (2) of Section~\ref{sec:arena}) is collected on a fixed anchored scale displayed on every annotation screen, so that scores are comparable across human evaluators, models, and levels. Human evaluators are shown the five anchors of Table~\ref{tab:rubric} and may assign any integer in $[0,10]$, interpolating between adjacent anchors. The same rubric is applied identically at all levels: it is the demonstration $V$, not a level label, that defines what a faithful imitation is, consistent with the level being withheld from the human evaluator.

\begin{table}[th]
\centering
\small
\caption{\textbf{Imitation-quality rubric.} The five fixed anchors shown to human evaluators for the per-rollout score $q \in [0,10]$ (question (2)). Intermediate integers interpolate between adjacent anchors.}
\label{tab:rubric}
\begin{tabular}{cl}
\toprule
Score & Anchor \\
\midrule
$0$  & Complete failure \\
$2$  & Minimal intent understanding \\
$5$  & Partial imitation, no adaptation \\
$8$  & Good imitation with minor errors \\
$10$ & Perfect imitation \\
\bottomrule
\end{tabular}
\end{table}

\paragraph{Pairwise-comparison protocol.} An A/B screen pairs two rollouts only when they are directly comparable, which we enforce by bucketing every evaluation rollout by its transfer setting and pairing only within a bucket. The three buckets are \emph{seen} (the five seen evaluation tasks), \emph{zero-shot} (the five unseen tasks executed directly from a pre-trained checkpoint), and \emph{transfer} (the same five unseen tasks after few-shot adaptation, comprising both the pre-train$+$fine-tune and the from-scratch conditions, which are therefore mutually comparable); the three settings are defined in Section~\ref{sec:setup}. Within a bucket, two rollouts are paired only when they share the same task, level, and episode, so the two videos differ only in the policy that produced them; rollouts from the same model at different pre-training scales ($15/30/45$) are comparable and are paired within the bucket. Simulation and real-world rollouts are never paired with each other, as they originate from distinct visual and dynamical domains; pairing is therefore confined to a single (domain, setting, task, level, episode) cell, and because the Arena randomizes which rollout is shown as A versus B, a pair is unordered. In simulation, where all baselines are evaluated, this yields $15{,}000$ annotated screens--$6{,}000$ seen, $6{,}000$ transfer, and $3{,}000$ zero-shot. On real hardware four models are evaluated; the $5{,}000$ real-world screens are therefore split $2{,}500$ seen and $2{,}500$ transfer.

\paragraph{Execution and scoring.} For each evaluation episode the platform rolls out the policy until the task terminates or a fixed horizon is reached, and records the full trajectory and the final scene state. Two scoring streams are then produced. The first is the per-rollout absolute-success judgment: in simulation, where the scene state is fully observable, it is scored by the final and sub-goal success rates defined below (fast, reproducible, and suited to large-scale evaluation) and \emph{additionally} by Arena human evaluation, so that the two can be cross-validated; on real hardware no automated metric is available and the Arena human judgment is the sole source. This judgment is the per-rollout task-completion question (question (1) of Section~\ref{sec:arena}), recorded on a three-way scale (\textsc{success}\,/\,\textsc{partial}\,/\,\textsc{fail}), together with the quality score $q$. The second stream is pairwise human preference, collected in both regimes as described above. The metrics built from these two streams are defined next.

\paragraph{Human success rate, mean quality, and win rate.} The human success rate is the fraction of rollouts judged \textsc{success}; a \textsc{partial} verdict does \emph{not} count toward success, matching the hard criterion of the automated $\mathrm{SR}$ below. Over a set of judged rollouts $\mathcal{R}$,
\[
\mathrm{SR}_{\text{human}} \;=\; \frac{1}{|\mathcal{R}|} \sum_{r \in \mathcal{R}}
  \mathbf{1}\!\bigl[\text{verdict}(r) = \textsc{success}\bigr],
\qquad
\overline{Q} \;=\; \frac{1}{|\mathcal{R}|} \sum_{r \in \mathcal{R}} q_r ,
\]
both computed over distinct rollouts ($36{,}000$ in simulation, $4{,}800$ on real hardware), so $\overline{Q}$ admits partial credit and separates failures that the binary success question collapses. For the pairwise preference question, each annotated screen awards $1$ to the preferred rollout, $0.5$ to each rollout on a tie, and $0$ to the other; a model's win rate $\mathrm{WR}$ is its mean awarded score over all screens in which it participates. Because pairing respects the (domain, setting, task, level) structure above, $\mathrm{SR}_{\text{human}}$, $\overline{Q}$, and $\mathrm{WR}$ can all be resolved by level \emph{post hoc} even though the level is hidden from human evaluators at judgment time.

\paragraph{Final success rate.} For each episode a binary success predicate checks whether all task-relevant goal states are satisfied at termination. Aggregated over the evaluation set $\mathcal{E}$,
\[
\mathrm{SR} \;=\; \frac{1}{|\mathcal{E}|} \sum_{e \in \mathcal{E}}
  \mathbf{1}\!\bigl[\text{all task-relevant goal-state predicates hold in } e\bigr].
\]
$\mathrm{SR}$ is a hard, binary criterion: an episode either satisfies all goal predicates or it does not. The phase definitions and success predicates are hand-crafted per task to reflect that task's goal states and intermediate milestones.

\paragraph{Sub-goal success rate.} Each task is decomposed into $M$ ordered sub-goal phases. For each phase $j$ the tracker records the peak value of the corresponding shaped sub-reward over the episode horizon, and the phase is counted as completed when that peak exceeds a fixed threshold. Letting $m_e \le M$ denote the number of completed phases in episode $e$,
\[
\mathrm{Sub\text{-}SR} \;=\; \frac{1}{|\mathcal{E}|}
  \sum_{e \in \mathcal{E}} \frac{m_e}{M}.
\]
$\mathrm{Sub\text{-}SR}$ measures the fraction of the task pipeline the policy executes and serves as a continuous proxy for imitation depth when $\mathrm{SR}$ is near zero.

\paragraph{Scope of the automated metrics.} Both metrics apply uniformly across all levels as general task-completion indicators, which is why the simulation experiments report them as a single overall score rather than per level; the per-level fidelity view is the responsibility of the Arena human evaluation. We do not define a trajectory-similarity metric because the human--robot embodiment gap leaves a trajectory-level predicate without an unambiguous reference, so $\mathrm{SR}$ and $\mathrm{Sub\text{-}SR}$ are the sole automated proxies. On real hardware neither automated metric is used, and all absolute success judgments come from Arena human evaluation.

\paragraph{Worked example: predicate and phases for \textsc{PourKettle}.} To make the hand-crafted predicates concrete, we give the full specification for one task. The \textsc{PourKettle} goal predicate is the conjunction of a fill condition and a spill condition,
\[
\text{success} \;=\; \bigl[\text{liquid in cup} \ge \tau_{\text{fill}}\bigr] \;\wedge\; \bigl[\text{liquid spilled} \le \tau_{\text{spill}}\bigr],
\]
with thresholds $\tau_{\text{fill}}=20$ and $\tau_{\text{spill}}=20$ on the simulated liquid volume. The task decomposes into $M=4$ ordered sub-goal phases--\emph{reach} (gripper within $\epsilon$ of the kettle handle), \emph{grasp} (stable contact, kettle lifted), \emph{transport} (kettle spout positioned above the cup), and \emph{pour} (kettle tilted past the pour angle while above the cup)--each with a shaped sub-reward whose peak is thresholded as in the $\mathrm{Sub\text{-}SR}$ definition.

\subsection{Baseline Adapter Details}
\label{app:baselines}

\begin{figure}[t]
\centering
\includegraphics[width=\textwidth]{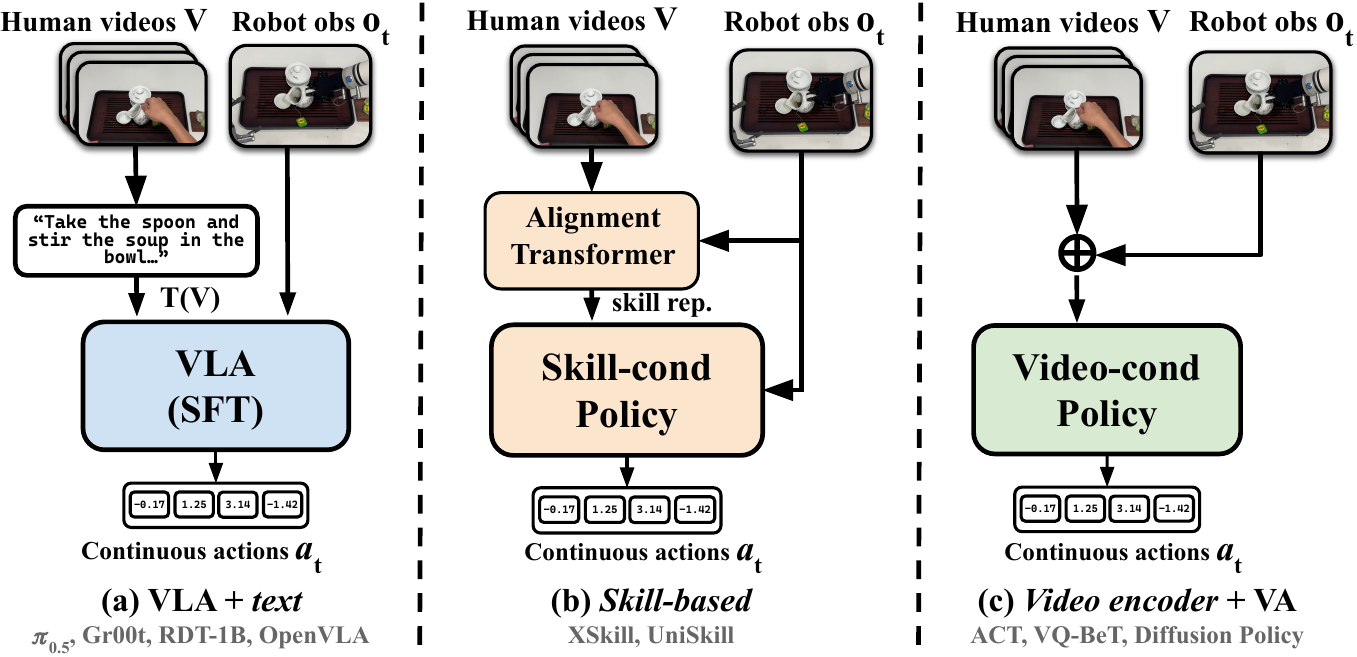}
\caption{\textbf{Adapting three paradigms to a unified control interface} $a_t = \pi(V, o_t)$. Across all families the backbone is frozen and only the action head is trained on IG-10K. \emph{(a) VLA + text:} a fixed captioner $T$ converts $V$ to a language prompt for a frozen vision--language backbone. \emph{(b) Skill-based:} a frozen encoder extracts a cross-embodiment skill representation from $V$, and a skill-conditioned policy acts on the retrieved skill index. \emph{(c) Video encoder + VA:} a frozen encoder maps $V$ to a task embedding concatenated with the observation tokens for ACT, Diffusion Policy, or VQ-BeT.}
\label{fig:baselines}
\end{figure}

This appendix expands Section~\ref{sec:baselines} with the per-family mechanisms, encoders, and training recipes. The video encoder or vision--language backbone is frozen and only the action-generation module is trained on IG-10K paired data, following each method's original recipe except where the imitator-game interface requires adaptation. Figure~\ref{fig:baselines} shows how each family is wired up to be called with the reference videos and robot observation; the full per-model hyperparameters are in Appendix~\ref{app:baseline-implementation}.

\paragraph{VLA family.} The four VLA backbones~\citep{bjorck2025gr00t,liu2025rdt,intelligence2025pi_,kim2024openvla} are language-conditioned, so the demonstration is first passed through the fixed captioner $T(V) \to \ell$, implemented via human annotation augmented with Qwen3-VL, which emits the per-arm sub-task sequence, the semantic task description, and the underlying intent--the same three abstraction levels carried as language annotation in IG-10K (Appendix~\ref{app:dataset}). The description $\ell$ is supplied as the language prompt and the action head is trained while the VLM backbone remains frozen. $T$ is held fixed across all VLA baselines so that performance differences reflect the action heads rather than the captioner; concretely, $T$ applies a single frozen prompt template to Qwen3-VL and emits the three abstraction levels in a fixed format, with no per-model tuning. A caption that was uniformly poor would pull every VLA down together, so we checked the caption channel itself in two ways. First, the captions carry the same information as the video: they state the task intent, the object layout, and the per-arm sub-task sequence together with its motions, and every caption is checked by a human annotator. Second, we test $T$ directly by replacing the VLM-augmented captions with the purely human-written ground-truth annotations of IG-10K and re-evaluating $\pi_{0.5}$: seen-task $\mathrm{SR}$ changes by $0.01 \pm 0.05$ across the three corpus scales, i.e.\ within noise. We did not search over caption styles, so it is still possible that shorter prompts suit small corpora better; a fuller ablation is future work.

\paragraph{Skill-based family.} XSkill~\citep{xu2023xskill} discovers a set of cross-embodiment skill prototypes from unlabeled human and robot videos via self-supervised contrastive learning with Sinkhorn--Knopp clustering, and conditions a diffusion policy on the prototype index retrieved for the current observation. UniSkill~\citep{kim2025uniskill} instead learns embodiment-agnostic skill representations through future-frame prediction (Inverse and Forward Skill Dynamics), which removes the need for domain alignment and allows skill transfer from human videos to robot policies trained solely on robot data. Both are trained from scratch on IG-10K, with the skill-conditioned policy stage trained with a frozen video extractor.

\paragraph{VA family.} The action-generation backbones~\citep{zhao2023learning,chi2025diffusion,lee2024behavior} are adapted by extracting a task representation from $V$ with one of three frozen video encoders--DINOv2-ViT-L/14~\citep{oquab2023dinov2}, SigLIP2-SO400M~\citep{tschannen2025siglip}, or VideoMAE-Large~\citep{tong2022videomae}--and concatenating the video embedding to the robot observation embedding at every inference step. The encoder is frozen and shared across all three backbones, and only the action head is trained from scratch, giving the three-encoder $\times$ three-head grid evaluated in Section~\ref{sec:experiments}.

\paragraph{Why the task encoder is frozen.} Freezing the video encoder or vision-language backbone is a comparability-versus-cost trade-off, and we tested the alternative before adopting it. In early runs we unfroze the task-related encoders -- the video encoder for the skill and VA families, the VLM backbone for the VLA family -- and observed comparable gains on both sides, giving no evidence that freezing penalizes the VLA family in particular. Unfreezing, however, forecloses caching: with a frozen task encoder the encoding of $V$ can be computed once and reused across epochs, which at the scale of our grid ($15$ trained variants $\times$ three corpus scales, plus few-shot regimes) is an orders-of-magnitude difference in training cost and is what makes the full comparison feasible. We therefore freeze the task encoder in every family and train only the action head, with comparable trainable-parameter counts, and treat a fully unfrozen comparison -- and a quantitative estimate of how much end-to-end adaptation adds per family -- as future work.

\paragraph{Training configuration.} All variants are trained on IG-10K under matched compute and a common optimizer schedule, with only the action-generation module updated. 

\section{Additional Experimental Results}
\label{app:exp}

\begin{figure}[th]
\centering
\includegraphics[width=\textwidth]{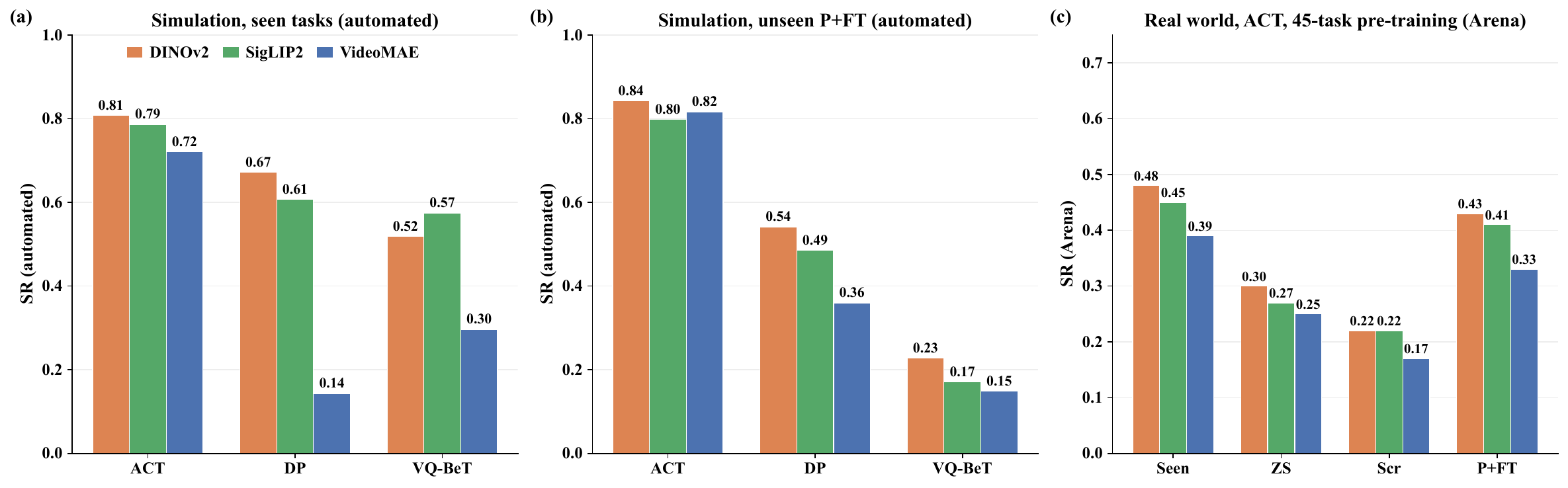}
\caption{\textbf{Q1: the frozen video encoder.} Success rate for the three Video-VA action heads
crossed with the three frozen encoders; the two simulation panels are the
automated channel and the real-world panel is the Arena channel.
Video encoders run with ACT on real-world hardware were only at the $45$-task pre-training scale.
}
\label{fig:c2-encoder}
\end{figure}

This appendix is organized around the three questions of Section~\ref{sec:experiments}, and
every figure answers exactly one of them.
\textbf{Q1 -- which imitation interface is strongest?}
Figure~\ref{fig:c1-paradigm} compares the three paradigms in simulation and under blind human
preference, and Figure~\ref{fig:c2-encoder} isolates the frozen video encoder inside the
Video-VA family.
\textbf{Q2 -- does scale help zero-shot transfer or few-shot adaptation?}
Figure~\ref{fig:c3-scaling} reports P+FT scaling and the per-model gain, and
Figure~\ref{fig:c4-levelscale} resolves the same experiment by hierarchy level.
\textbf{Q3 -- where does the hierarchy become hard?}
Figure~\ref{fig:c5-perlevel} contrasts the clean real-world coarse-to-fine decline with the
non-monotone simulation profile, and Table~\ref{tab:demo-swap} checks that the
demonstration is what specifies the task in the first place.
Figure~\ref{fig:c7-validity} then asks how far the automated and human scores,
and the simulated and physical domains, can stand in for one another; the per-task numbers behind
all of these aggregates are in Table~\ref{tab:per-task-reference}.
Two scoring channels appear throughout and are always named. The
\emph{automated} channel is the simulated success predicate, reported as $\mathrm{SR}$ and
$\mathrm{Sub\text{-}SR}$; the \emph{Arena} channel is human judgment, reported as
$\mathrm{SR}_{\mathrm{human}}$, the imitation score $\overline{Q}$ on the $0$--$10$ rubric of
Appendix~\ref{app:arena}, and the blind A/B win rate $\mathrm{WR}$. The Arena runs on both
domains: on the simulated rollouts, where it can be compared against the automated channel
directly, and on hardware, where there is no automated predicate and every reported
$\mathrm{SR}$ is therefore a human verdict.
Throughout, simulation numbers are averaged over the $\{15,30,45\}$ corpus scales as in the
body, and the Video-VA family is shown at its DINOv2 backbone unless noted.

\subsection{Q1: which imitation interface is strongest?}
\label{app:q1-paradigm}

\begin{figure}[t]
\centering
\includegraphics[width=\textwidth]{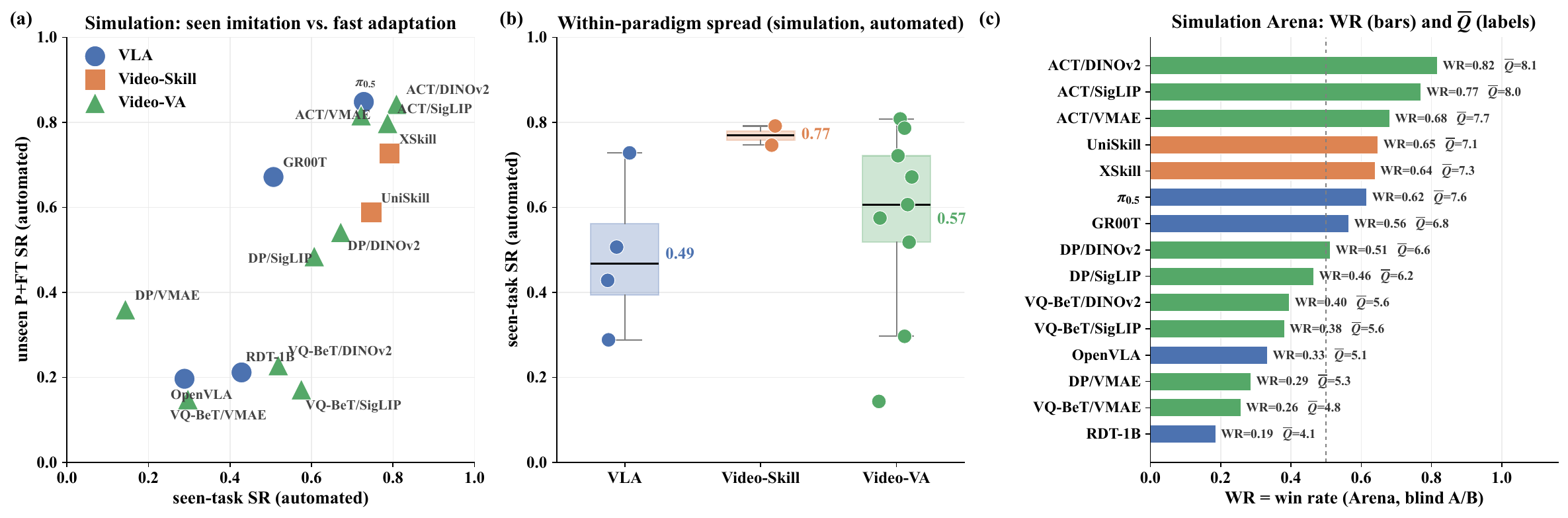}
\caption{\textbf{Q1: which imitation interface is strongest.}
\textbf{(a)} Each trained variant on the $(\mathrm{seen\text{-}SR},\,\mathrm{P{+}FT\text{-}SR})$
plane, automated channel; the video families cluster upper-right while the VLA family stretches
along the diagonal.
\textbf{(b)} Distribution of seen-task $\mathrm{SR}$ (automated) within each paradigm; the
within-family spread exceeds the widest between-family gap for VLA and Video-VA, and Video-Skill
is the tightest.
\textbf{(c)} Simulation Arena, human channel. Bar length is the blind A/B win
rate $\mathrm{WR}$; the label beside each bar gives that variant's $\mathrm{WR}$ together with its
mean imitation score $\overline{Q}$, which is a separate quantity and is not plotted. The human
channel reproduces the ranking that the automated channel gives in panels (a,b) on the same
rollouts.}
\label{fig:c1-paradigm}
\end{figure}

Figure~\ref{fig:c1-paradigm} unpacks the headline ranking of Section~\ref{sec:exp-video-language}
along three axes. Panel~(a) places all fifteen trained variants on the
$(\mathrm{seen\text{-}SR},\,\mathrm{P{+}FT\text{-}SR})$ plane, both axes automated. The two
video-conditioned families cluster in the upper right, while the VLA family stretches along the
diagonal from OpenVLA ($0.29$) to $\pi_{0.5}$ ($0.73$); seen-task strength and few-shot
adaptability are correlated but not interchangeable, and the variants furthest above the diagonal
(GR00T, $\pi_{0.5}$) adapt better than their seen-task score would predict.
Panel~(b) shows why a family-level bar chart would mislead. Inside the VLA family
the variants span $0.29$--$0.73$ and inside Video-VA they span $0.14$--$0.81$, so for two of the
three families the spread within the family is wider than the widest gap between any two family
means (family means $0.49$, $0.77$, $0.57$; widest gap $0.28$). Video-Skill is by far the
tightest: its two members behave alike, which is what makes it predictable, not necessarily what
makes it best.
Panel~(c) shows that the ordering is not an accident of how the automated
predicate is written. Under blind A/B judgment in the \emph{simulation} Arena -- the same
rollouts, scored by people instead of by the predicate -- win rate gives the same family ranking
(Video-Skill $0.64$, Video-VA $0.50$, VLA $0.42$), with ACT/DINOv2 the single most preferred
variant ($\mathrm{WR}=0.82$, $\overline{Q}=8.1$).

\subsection{Q1: does the frozen video encoder matter?}
\label{app:q1-encoder}
 
Because every Video-VA variant receives its task specification through a frozen encoder, the
encoder is a design choice of for its action head. Figure~\ref{fig:c2-encoder}
separates the two.
On seen tasks (panel~a, automated) the encoder ordering for ACT is DINOv2 $\ge$
SigLIP2 $>$ VideoMAE ($0.81/0.79/0.72$), but the size of the effect depends on which action head
it is paired with: VideoMAE costs ACT $0.09$ SR and costs Diffusion Policy $0.53$
($0.67 \to 0.14$), which is close to total failure. 
Encoder and action head therefore cannot be
chosen independently.
After few-shot adaptation (panel~b, automated), the gap almost closes for ACT
($0.84/0.80/0.82$) but not for DP or VQ-BeT: ten demonstrations can make up for a poor task
embedding only when the action head is already strong.
Panel~(c) replicates the ablation on hardware, where the scores are Arena
human verdicts. DINOv2, SigLIP2, and VideoMAE were run on hardware with ACT only at the
$45$-task pre-training scale. At this matched scale the
hardware ordering reproduces the simulation ordering in every regime: VideoMAE is last
everywhere, and the two image-level encoders are close to each other, align with the simulation results.

\subsection{Q2: pre-training scale, zero-shot transfer, and few-shot adaptation}
\label{app:q2-scaling}

\begin{figure}[t]
\centering
\includegraphics[width=\textwidth]{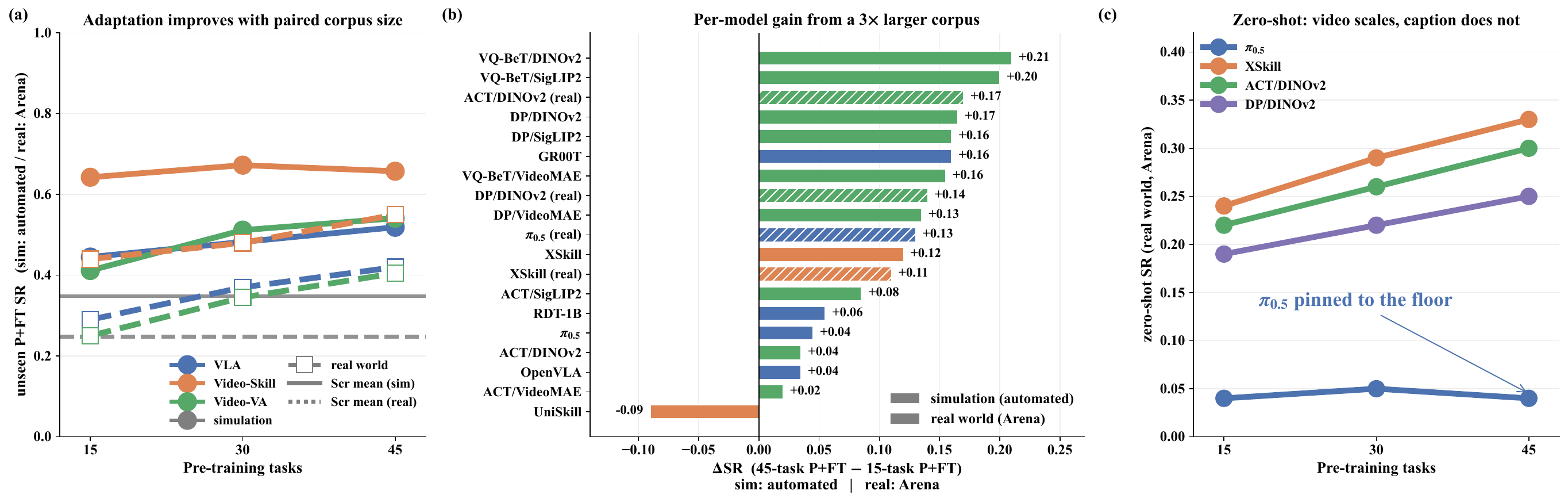}
\caption{\textbf{Q2: scaling the paired pre-training corpus.}
\textbf{(a)} P+FT $\mathrm{SR}$ vs.\ corpus size, simulation (solid, automated) and real world
(dashed, Arena); the two grey horizontal lines are the \textsc{scratch} mean for
each domain -- simulation over all $15$ variants, real world over the four
representative models -- computed on the same basis as the \textsc{Scr.}\ columns of
Table~\ref{tab:main-sim} and Table~\ref{tab:main-real}.
\textbf{(b)} Per-model $\Delta\mathrm{SR}$ from $15$ to $45$ pre-training tasks; real-world
models are hatched and are Arena numbers. Positive for $18$ of $19$ variants.
\textbf{(c)} Real-world unseen zero-shot (Arena) vs.\ scale: the three video-conditioned models
improve monotonically while the caption-conditioned $\pi_{0.5}$ stays at the floor.}
\label{fig:c3-scaling}
\end{figure}

Figure~\ref{fig:c3-scaling} is the evidence behind the scaling claim of
Section~\ref{sec:exp-scaling}.
Panel~(a) plots P+FT success against corpus size in both domains -- automated in
simulation, Arena on hardware. All three paradigms improve, in simulation and on hardware alike,
and by $45$ tasks every family's P+FT curve lies above the \textsc{scratch} baseline for its own
domain (sim $0.35$, averaged over all $15$ variants; real $0.25$, averaged over the four
representative models). 
The gain therefore comes from the paired
corpus, not from the ten few-shot demonstrations, which both conditions receive.
Panel~(b) resolves this per model. Going from $15$ to $45$ pre-training tasks
raises $\Delta\mathrm{SR}$ for $18$ of the $19$ trained variants across the two domains
($14/15$ in simulation, $4/4$ on hardware). We read this as a
limit of its skill-dynamics stage, though with one data point we cannot separate that from noise.
Panel~(c) reports what the aggregate hides. On hardware, zero-shot success on
unseen tasks rises steadily with corpus size for all three video-conditioned models
(XSkill $0.24 \to 0.33$, ACT/DINOv2 $0.22 \to 0.30$, DP/DINOv2 $0.19 \to 0.25$), but stays at
about $0.04$ for $\pi_{0.5}$ at every scale. More paired human-robot video therefore buys
zero-shot ability only for policies that watch the video itself; passing the same demonstration
through a caption first does not turn extra pre-training into transfer.

\subsection{Q2: is the scaling benefit confined to the specific levels?}
\label{app:q2-levelscale}

\begin{figure}[t]
\centering
\includegraphics[width=\textwidth]{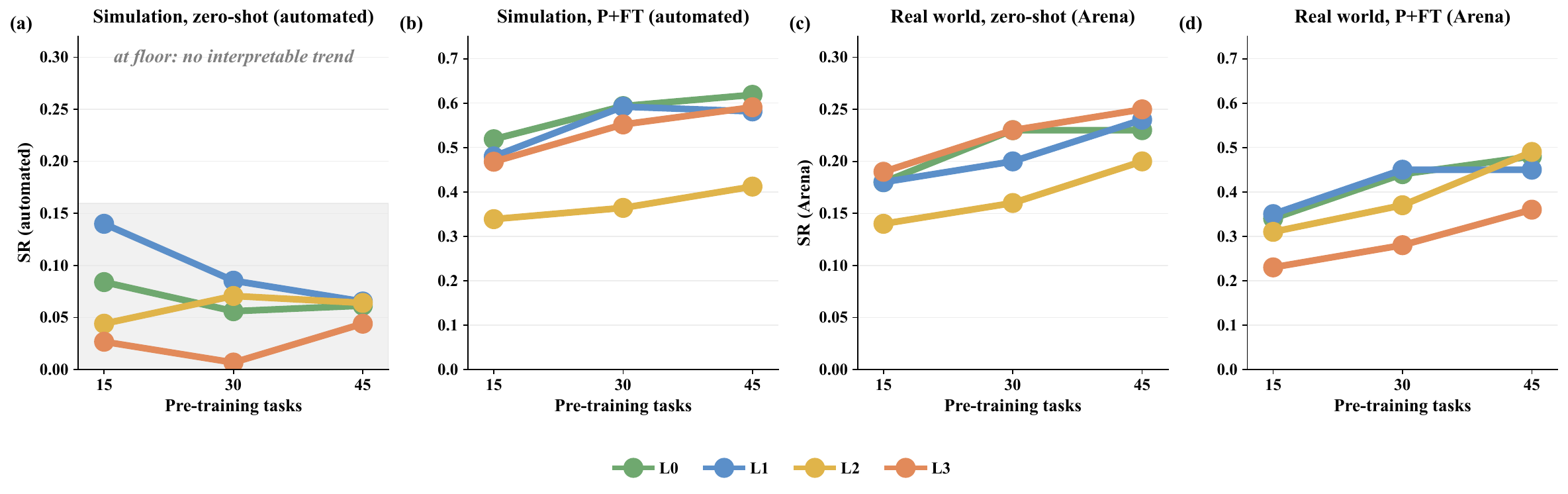}
\caption{\textbf{Q2: scaling resolved by hierarchy level.} Unseen-task success against
the number of pre-training tasks, separately per level, for
\textbf{(a)} simulation zero-shot, \textbf{(b)} simulation P+FT, \textbf{(c)} real-world
zero-shot and \textbf{(d)} real-world P+FT. Simulation curves average all $15$ trained variants;
real-world curves average the four representative models. Note the different $y$ ranges for
each columns.}
\label{fig:c4-levelscale}
\end{figure}

The body averages over corpus scales for compactness. Figure~\ref{fig:c4-levelscale} resolves the zero-shot and
P+FT settings by level \emph{and} scale in both domains, and Table~\ref{tab:level-scale} gives the
same numbers. Three readings follow.
First, in every panel except simulation zero-shot, success improves monotonically or
near-monotonically from $15$ to $45$ pre-training tasks \emph{at every level}: the benefit is not
confined to the easy end of the hierarchy.
Second, simulation zero-shot (panel~a) sits at the floor throughout
($\mathrm{SR} \le 0.14$) and its level ordering moves around without a readable pattern. We draw
no conclusion from it, and do not use it to rank encoders or paradigms.
Third, the L3 curve behaves differently in the two domains. In the real world (panel~d) it
improves with scale ($0.23 \to 0.36$) while remaining a clear band below L0--L2 at every scale,
i.e.\ more paired pre-training helps at the intent level but does not close the intent-level gap.
In simulation (panel~b) L2 rather than L3 is the depressed level, for the construction reason
analyzed in Section~\ref{app:q3-perlevel}. 

\begin{table}[th]
\centering
\small
\setlength{\tabcolsep}{5pt}
\caption{\textbf{Success rate by hierarchy level and pre-training scale.} Unseen-task zero-shot (ZS) and pre-train$+$fine-tune (P+FT) success, resolved by level and by the number of pre-training tasks ($15/30/45$). Simulation numbers average all $15$ trained variants; real-world numbers average the four representative models. Simulation columns are the automated channel, real-world columns the Arena channel. These are the numbers plotted in Figure~\ref{fig:c4-levelscale}.}
\label{tab:level-scale}
\begin{tabular}{lccccccccccccc}
\toprule
& \multicolumn{3}{c}{Sim ZS (auto.)} & \multicolumn{3}{c}{Sim P+FT (auto.)} & \multicolumn{3}{c}{Real ZS (Arena)} & \multicolumn{3}{c}{Real P+FT (Arena)} \\
\cmidrule(lr){2-4}\cmidrule(lr){5-7}\cmidrule(lr){8-10}\cmidrule(lr){11-13}
Level & 15 & 30 & 45 & 15 & 30 & 45 & 15 & 30 & 45 & 15 & 30 & 45 \\
\midrule
L0 & 0.08 & 0.06 & 0.06 & 0.52 & 0.59 & 0.62 & 0.18 & 0.23 & 0.23 & 0.34 & 0.44 & 0.48 \\
L1 & 0.14 & 0.09 & 0.07 & 0.48 & 0.59 & 0.58 & 0.18 & 0.20 & 0.24 & 0.35 & 0.45 & 0.45 \\
L2 & 0.04 & 0.07 & 0.06 & 0.34 & 0.36 & 0.41 & 0.14 & 0.16 & 0.20 & 0.31 & 0.37 & 0.49 \\
L3 & 0.03 & 0.01 & 0.04 & 0.47 & 0.55 & 0.59 & 0.19 & 0.23 & 0.25 & 0.23 & 0.28 & 0.36 \\
\bottomrule
\end{tabular}
\end{table}

\subsection{Q3: where the hierarchy becomes hard}
\label{app:q3-perlevel}

\begin{figure}[t]
\centering
\includegraphics[width=\textwidth]{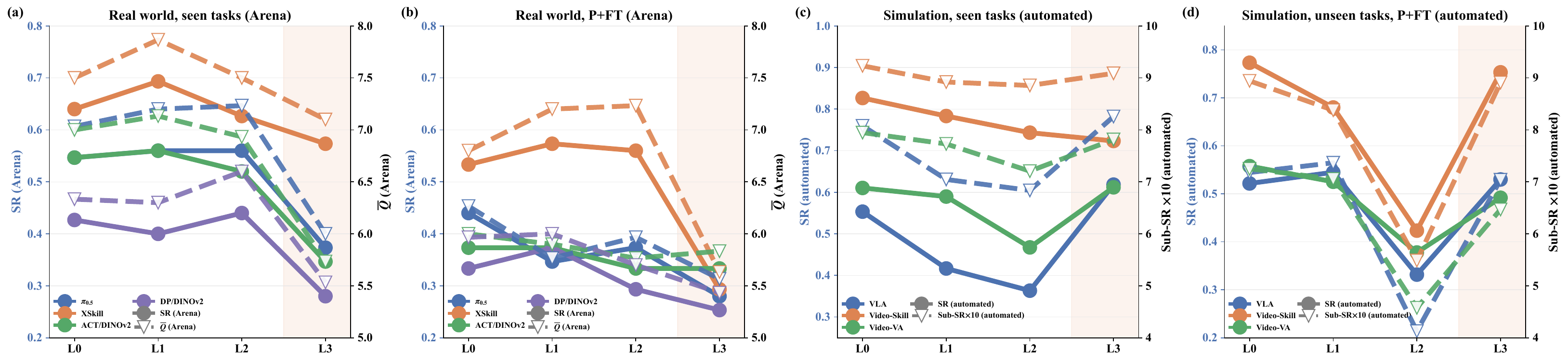}
\caption{\textbf{Q3: per-level success and completion quality.}
\textbf{(a)} Real-world seen-task $\mathrm{SR}$ (Arena, left axis, solid) across L0--L3, one curve
per representative model; each model's imitation score $\overline{Q}$ (right axis, dashed, same
color, also Arena) is overlaid.
\textbf{(b)} Real-world P+FT $\mathrm{SR}$ and $\overline{Q}$ (Arena), same layout as~(a).
\textbf{(c)} Simulation seen-task $\mathrm{SR}$ (automated, left axis, solid) by paradigm family;
each family's Sub-SR$\times10$ (right axis, dashed, same color, also automated) is overlaid.
\textbf{(d)} Simulation unseen-task P+FT $\mathrm{SR}$ and Sub-SR$\times10$ (automated), same
layout as~(c).}
\label{fig:c5-perlevel}
\end{figure}

Figure~\ref{fig:c5-perlevel} lays out four panels in a single row -- real world (Seen, P+FT) then
simulation (Seen, P+FT) -- success rate on the primary axis and a completion-quality score on the
secondary axis in every panel.
In panel~(a), success is flat from L0 to L2
($0.54/0.55/0.54$) and drops at L3 ($0.39$); $\overline{Q}$ follows the same shape ($6.97/7.12/
7.07 \to 6.09$). XSkill is strongest at every level ($0.64/0.69/0.63/0.57$) and still loses ground
at L3.
In panel~(b), success is flat from L0 to L2
($0.42/0.42/0.39$) and drops at L3 ($0.29$). XSkill carries the drop
($\mathrm{SR}\approx0.53$--$0.57$ through L2, falling to $0.29$ at L3; $\overline{Q}$ $7.2\to5.6$),
while the other three models sit in a comparatively narrow $\overline{Q}\approx5.4$--$6.3$ band at
every level. Current policies cope with a rearranged layout, or a replacement object of the same
kind, but not with an object that has to be handled differently.
In panel~(c), the all-model curve is non-monotone: it
dips at L2 and rebounds at L3 ($0.62/0.57/0.48/0.63$), and the VLA family peaks at L3 above its
own L0 value ($0.55/0.42/0.36/0.62$). The per-family Sub-SR curves show the identical
dip-then-rebound shape (VLA $0.81/0.70/0.68/0.83$, Video-Skill $0.92/0.89/0.89/0.91$, Video-VA
$0.79/0.77/0.72/0.78$), so the effect is not confined to one paradigm or to the binary predicate.
Figure~\ref{fig:c7-validity}(b) reproduces this shape on the same rollouts under Arena scoring, so
it is not an artifact of the automated channel.
In panel~(d), the L2 dip recurs
($0.58/0.55\to0.37$) with a partial rebound at L3 ($0.54$) that stays below L0, unlike panel~(c).
The pattern holds by family (VLA $0.52/0.55/0.33/0.53$, Video-Skill $0.77/0.68/0.42/0.75$,
Video-VA $0.56/0.53/0.38/0.49$) and in Sub-SR (VLA $0.72/0.74/0.41/0.70$, Video-Skill
$0.89/0.84/0.55/0.89$, Video-VA $0.73/0.71/0.46/0.65$), so the L2 dip is a property of the
simulated L2 condition itself rather than of the seen/unseen split.

\textbf{Why L2 dips in simulation.} Two construction choices explain it. First, the simulated L2
condition also swaps which arm performs the manipulation -- deliberately, to remove any
motion-level shortcut, since L1 already moves the objects and L3 already changes what has to be
done. Second, the simulation asset pool is finite and its
motions are rule-based, so on simple pick-and-place tasks a substituted L3 object can often still
be handled with roughly the strategy seen in training. Hardware offers no such shortcut: a
functionally different object differs in mass, shape and how it must be gripped, and the robot has
to adapt for real. This is why the body reports the simulation aggregate without splitting it by
level, and reads the per-level picture from the Arena instead.

\begin{table}[th]
\centering
\small
\setlength{\tabcolsep}{3.4pt}
\caption{\textbf{Demonstration-swap sanity check (simulation, seen tasks, $45$-task checkpoints, automated channel).} Success rate when the conditioning demonstration is the correct one (\emph{original}) versus a video of a similar or an unrelated task, with the imitator scene unchanged.}
\label{tab:demo-swap}
\begin{tabular}{lccccc ccccc}
\toprule
& \multicolumn{5}{c}{ACT/DINOv2} & \multicolumn{5}{c}{DP/DINOv2} \\
\cmidrule(lr){2-6}\cmidrule(lr){7-11}
Demo video & L0 & L1 & L2 & L3 & Avg & L0 & L1 & L2 & L3 & Avg \\
\midrule
original  & 0.87 & 0.77 & 0.72 & 0.88 & 0.81 & 0.72 & 0.66 & 0.59 & 0.71 & 0.67 \\
similar   & 0.32 & 0.29 & 0.32 & 0.32 & 0.31 & 0.02 & 0.05 & 0.01 & 0.04 & 0.03 \\
unrelated & 0.34 & 0.28 & 0.39 & 0.19 & 0.30 & 0.00 & 0.00 & 0.00 & 0.00 & 0.00 \\
\midrule
& \multicolumn{5}{c}{$\pi_{0.5}$} & \multicolumn{5}{c}{XSkill} \\
\cmidrule(lr){2-6}\cmidrule(lr){7-11}
Demo video & L0 & L1 & L2 & L3 & Avg & L0 & L1 & L2 & L3 & Avg \\
\midrule
original  & 0.85 & 0.67 & 0.70 & 0.69 & 0.73 & 0.87 & 0.85 & 0.77 & 0.68 & 0.79 \\
similar   & 0.09 & 0.18 & 0.09 & 0.23 & 0.15 & 0.47 & 0.41 & 0.05 & 0.29 & 0.31 \\
unrelated & 0.01 & 0.00 & 0.00 & 0.02 & 0.01 & 0.39 & 0.23 & 0.03 & 0.27 & 0.23 \\
\bottomrule
\end{tabular}
\end{table}

\subsection{Q3: is the demonstration actually used?}
\label{app:demo-swap}

A policy that had simply memorized its training tasks could reproduce much of
Section~\ref{sec:exp-video-language} without reading the conditioning video at all, since the
imitator scene alone often identifies the task. We test this on the five seen tasks with the
$45$-task checkpoints: at inference we replace the conditioning demonstration $V$ with a video of
a \emph{different} IG-10K task, leaving the simulated imitator scene, the robot and every other
input untouched, and re-run the evaluation at all four levels. In the \emph{similar} regime the
replacement shares broad manipulation structure with the original
(\textsc{StirSpoon}$\to$\textsc{GrindFood}, \textsc{FoldTowel}$\to$\textsc{PlaceClothBasket},
\textsc{PlaceMugRack}$\to$\textsc{PlaceCommodityRack},
\textsc{PlaceFileFolder}$\to$\textsc{PlaceMagazineFolder},
\textsc{PlacePlateRack}$\to$\textsc{PlaceBookBookcase}); in the \emph{unrelated} regime it shares
nothing (\textsc{OpenBox}, \textsc{PickWash}, \textsc{PlaceBrushRest},
\textsc{PlaceScrewdriver}, \textsc{ScanPillBottle}, assigned in that order).

\subsection{How far do the two evaluation channels, and the two domains, agree?}
\label{app:validity}

\begin{figure}[t]
\centering
\includegraphics[width=\textwidth]{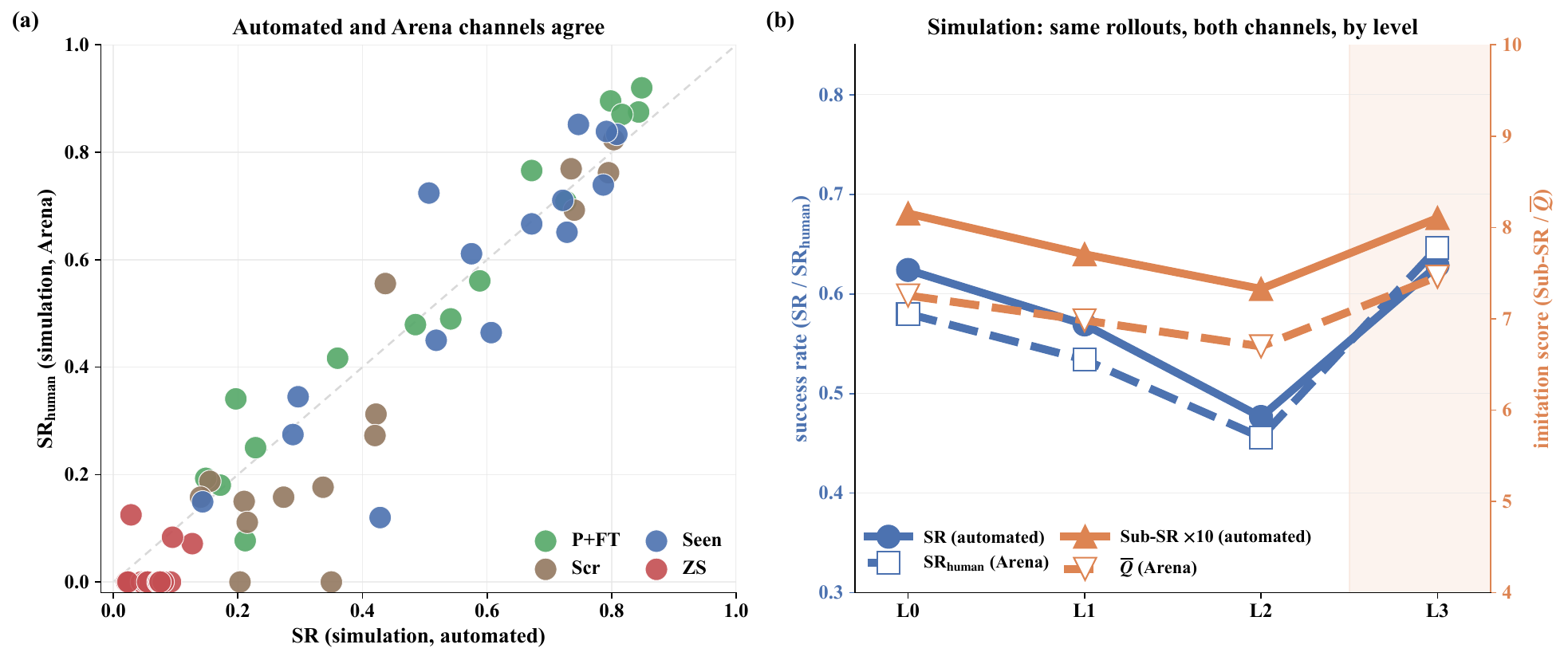}
\caption{\textbf{Validity of the evaluation channels.}
\textbf{(a)} Automated vs.\ Arena scoring of the identical simulation rollouts
($x$: automated $\mathrm{SR}$; $y$: Arena $\mathrm{SR}_{\mathrm{human}}$). Dashed line is $y=x$.
\textbf{(b)} The same two channels, resolved by hierarchy level on the seen-task rollouts (left
axis: automated $\mathrm{SR}$ and Arena $\mathrm{SR}_{\mathrm{human}}$; right axis: automated
Sub-SR$\times10$ and Arena $\overline{Q}$).}
\label{fig:c7-validity}
\end{figure}

The benchmark reports two score streams and two domains, and Figure~\ref{fig:c7-validity}
quantifies how far each can substitute for the other.
Panel~(a) plots the automated predicate against the Arena judgment on the \emph{identical}
simulation rollouts. 
The two agree
closely: $r = 0.858$ for $\mathrm{SR}$ vs.\
$\mathrm{SR}_{\mathrm{human}}$ and $r = 0.861$ for $\mathrm{Sub\text{-}SR}$ vs.\ $\overline{Q}$ when the same cells are resolved by level. The automated metric is
therefore a faithful, cheap proxy for the human judgment in simulation, which is what licenses
running the full $15$-variant grid automatically.
Panel~(b) resolves the same agreement by hierarchy level, on the seen-task rollouts: the automated $\mathrm{SR}$ and Arena $\mathrm{SR}_{\mathrm{human}}$
curves, and separately the automated Sub-SR and Arena $\overline{Q}$ curves, both dip at L2 and
both reproduce the L3 rebound (Section~\ref{app:q3-perlevel}).

\subsection{Per-task diagnostics}
\label{app:pertask}

Beyond the ten seen/unseen tasks used for cross-model comparison in the main
text, we also report a full-coverage sanity check: DP and ACT trained on the complete IG-10K
simulation corpus (all $50$ tasks, $50$ demonstrations each, all $200$ task--level variants) and
evaluated on every task. The purpose is not to compare these two models against the others, but to
confirm that the ten selected tasks are representative of the wider pool -- i.e., that the
benchmark's difficulty comes from the L0--L3 hierarchy itself rather than from an unrepresentative
or favorably chosen task subset -- and to leave future work a reference that is not limited to the
ten tasks used elsewhere in the paper. The full per-task reference is Table~\ref{tab:per-task-reference} in
Appendix~\ref{app:per-task-reference}.

\section{Mask Generation Details}
\label{app:mask-generation}

The segmentation masks in IG-10K are produced as an offline preprocessing step rather than as part of the policy input or the benchmark scoring pipeline. They serve as an additional object-centric annotation layer of the dataset for analysis, dataset inspection, and future training objectives.

\paragraph{Text-conditioned mask proposal.}
We use a Grounded-SAM-2 based pipeline to generate the masks. For each task, annotators manually inspect episodes and summarize the relevant objects for human, robot, and simulation. The object list is converted into text prompts for GroundingDINO, which proposes object boxes for each camera episode using task-specific box and text confidence thresholds. These boxes are then passed to SAM-2 to produce the corresponding video masks for that episode. The procedure therefore combines human-curated object prompts with open-vocabulary detection and video segmentation.

\paragraph{Filtering and quality control.}
GroundingDINO detections are filtered by box and text confidence thresholds. In our runs we use conservative defaults of \texttt{box\_threshold}=0.35 and \texttt{text\_threshold}=0.25. We additionally render mask overlays on sampled camera episodes for visual inspection, run audit scripts to detect empty or missing episode-camera masks, and manually filter problematic results before treating the masks as part of the released annotation set.

\section{MANO Hand Pose Annotation}
\label{app:mano-processing}

MANO hand pose annotations in IG-10K are auto-generated and serve as auxiliary hand annotations.

\paragraph{Per-view WiLoR estimation.}
We estimate MANO-style hand poses with WiLoR on multi-view RGB videos. The camera streams are synchronized at the episode timestep level. Each view is processed independently, and predictions are stored separately for the left and right hands.

For frame $t$, camera $c$, and hand side $h \in \{\mathrm{left}, \mathrm{right}\}$, we store
\[
\theta^{g}_{t,c,h} \in \mathbb{R}^{3}, \quad
\theta^{p}_{t,c,h} \in \mathbb{R}^{45}, \quad
\beta_{t,c,h} \in \mathbb{R}^{10},
\]
where $\theta^{g}$ is the MANO global orientation, $\theta^{p}$ contains the 15 articulated MANO joint rotations, and $\beta$ denotes MANO shape coefficients. Rotations are stored in axis-angle form. We additionally store
\[
\tau_{t,c,h} \in \mathbb{R}^{3}, \quad
f_{t,c,h} \in \mathbb{R}^{2}, \quad
J_{t,c,h} \in \mathbb{R}^{21 \times 3},
\]
where $\tau$ is the full-image camera translation converted from WiLoR's crop-level camera prediction, $f$ is the focal length used for projection, and $J$ denotes the 21 predicted 3D hand keypoints. Missing detections are stored as zero-valued placeholders with an invalid flag. Since calibrated camera extrinsics are unavailable, we do not transform estimates into a shared world coordinate frame; the annotations should be interpreted as per-view MANO estimates.

\paragraph{Shape-stabilized keypoints.}
To reduce temporal jitter caused by frame-wise MANO shape estimation, we compute an episode-level mean MANO shape $\bar{\beta}_{c,h}$ from valid detections and use it only to recompute the stored 3D keypoints,
\[
\bar{J}_{t,c,h}
=
\mathrm{MANO}_{J}
\left(
\theta^{g}_{t,c,h},
\theta^{p}_{t,c,h},
\bar{\beta}_{c,h}
\right),
\]
while keeping other MANO parameters unchanged.

\section{Baseline Implementation Details}
\label{app:baseline-implementation}

All baselines use the same IG-10K task splits, ZED2i RGB stream, and 16-dimensional dual-arm joint-position/gripper action space. For unseen-task transfer, both scratch and pre-train$+$fine-tune use the same 10 robot demonstrations per task.

\subsection{UniSkill}
\label{app:uniskill-implementation}

We keep the two-stage UniSkill pipeline: skill-dynamics learning followed by a skill-conditioned diffusion policy. To align different human and robot execution speeds, the robot-side IDM transition uses \texttt{robot\_frame\_gap}=35. Stage 1 trains the UniSkill dynamics modules with 10-frame, 30 Hz clips, batch size 192, AdamW, learning rate $1{\times}10^{-5}$, cosine decay, 500 warmup steps, and bfloat16 precision, while the VAE, text encoder, and depth estimator remain frozen. Stage 2 trains the policy with observation horizon 2, prediction horizon 16, action horizon 8, batch size 128, AdamW, learning rate $1{\times}10^{-4}$, DDPM diffusion with 100 steps, and alignment loss weight 1.0.

\subsection{XSkill}
\label{app:xskill-implementation}

We preserve XSkill's two-stage prototype-learning and prototype-conditioned policy structure, replacing offline prototype labels with prototypes extracted from paired IG-10K clips. Stage 1 uses 30-frame clips split by \texttt{slide}=3 into 4-frame windows, 128 prototypes, batch size 128, Adam, initial learning rate $2{\times}10^{-4}$, OneCycleLR with maximum learning rate $4{\times}10^{-4}$, temperature 0.1, and 3 Sinkhorn iterations. Stage 2 snaps each human video to 30 prototype tokens and trains a DDPM policy with observation horizon 4, prediction horizon 16, action horizon 8, batch size 128, AdamW, learning rate $1{\times}10^{-4}$, cosine decay, 500 warmup steps, and 60 diffusion steps. At evaluation time, the robot-side prototype is computed from 4 frames sampled from the recent observation history.

\subsection{OpenVLA}
\label{app:openvla-implementation}

We initialize from \texttt{openvla/openvla-7b} and keep the 7B vision-language backbone frozen. The human video is converted into the fixed caption $T(V)$, and only the OFT-style~\citep{kim2025fine} continuous action head and proprioceptive projector are trained for our 16-DoF dual-arm action space. To reduce training cost, we precompute action-token hidden states with shape $(256,4096)$ per sample and add proprioception through an output-side projector before the action head. The action horizon is 16; training uses AdamW, learning rate $1{\times}10^{-4}$, bfloat16 autocast, batch size 256, and 10 epochs across all three pre-training scales (45/30/15 tasks). Transfer runs also use 10 epochs with batch size 512.

\subsection[Pi0.5 implementation details]{$\pi_{0.5}$ implementation details}
\label{app:pi05-implementation}

We adapt the public $\pi_{0.5}$-DROID checkpoint. The PaliGemma vision-language prefix remains frozen, and the trainable modules are the action-expert branch, action projections, and time-conditioning MLPs. Inputs are the ZED2i RGB frame resized to $224{\times}224$, normalized dual-arm state, and the fixed caption $T(V)$. The model predicts 50-step continuous action chunks and is trained with the standard flow-matching objective. All pretraining runs use 10 epochs, AdamW, peak learning rate $3{\times}10^{-5}$ with cosine decay to $3{\times}10^{-6}$, 1000 warmup steps, batch size 16, bfloat16 precision, and 10 denoising steps at inference.

\subsection{RDT-1B implementation details}
\label{app:rdt-implementation}

We initialize from the public RDT-1B checkpoint and use the fixed caption $T(V)$ as language input. SigLIP-SO400M/14 and T5-v1.1-XXL are frozen, with text features cached when possible. We train LoRA adapters in the RDT Transformer and condition adaptors for language, image, state, and action tokens; the base RDT weights remain frozen. Actions are mapped into the original RDT unified action space with a validity mask, using a 16-step horizon and 30 Hz control-frequency token. Training uses LoRA rank 576, alpha 1152, dropout 0.05, batch size 128, gradient accumulation 4, AdamW, learning rate $1{\times}10^{-4}$, cosine schedule with 500 warmup steps, and bfloat16 precision. All pretraining runs use 10 epochs.

\subsection{GR00T implementation details}
\label{app:gr00t-implementation}

We adapt GR00T-N1.6-3B with the Eagle vision-language backbone frozen. The fixed caption $T(V)$, current ZED2i RGB frame, and dual-arm proprioception condition a 16-step action chunk. We train the flow-matching DiT action expert together with embodiment-specific state/action encoders, decoder, action position embedding, and action-side normalization layers. For all pretraining runs, GR00T is trained for 10 epochs with AdamW, learning rate $1{\times}10^{-4}$, weight decay $1{\times}10^{-5}$, cosine schedule with 0.05 warmup ratio, batch size 256, bfloat16 precision, and 4 denoising steps at inference. Dataset statistics are recomputed for the \texttt{NEW\_EMBODIMENT} normalization before each run.

\subsection{Video-VA (ACT, Diffusion Policy, VQ-BeT) Implementation Details}
\label{app:va-implementation}

ACT, Diffusion Policy, and VQ-BeT are conditioned on a cached task embedding $z \in \mathbb{R}^{256}$ extracted from four uniformly sampled frames of the human video. We evaluate DINOv2-ViT-L/14, SigLIP2-SO400M, and VideoMAE-Large as frozen encoders, with a one-layer adapter trained with the policy. ACT injects $z$ as an additional Transformer decoder token and uses L1 reconstruction plus KL loss ($\lambda_{\mathrm{KL}}=10$), action horizon 24, batch size 256, AdamW, learning rate $1{\times}10^{-4}$, and 100 epochs. Diffusion Policy concatenates $z$ into the global FiLM conditioning vector and uses a DDPM 1D UNet with action horizon 16, 100 train/inference diffusion steps, batch size 256, AdamW, learning rate $1{\times}10^{-4}$, and 100 epochs. VQ-BeT prepends a projected $z$ token to the observation sequence; its VQ-VAE is trained for 100 epochs on robot action chunks, then the GPT policy is trained for 100 epochs with batch size 256 and learning rate $1{\times}10^{-4}$.

\paragraph{Evaluation.}
At inference, each video-conditioned policy encodes $V$ once at episode start, while VLA policies reuse the fixed caption $T(V)$. Simulation uses the rollout horizons and predicates in Appendix~\ref{app:arena}, with $10$ trials per task--level pair across all $15$ variants; real-world evaluation uses $\pi_{0.5}$, XSkill, ACT/DINOv2, and DP/DINOv2 with 5 trials per task--level pair. Both domains evaluate the same $40$ task--level pairs (Appendix~\ref{app:splits}).

\section{Failure Modes and Task-Level Diagnostics}
\label{app:case-study}

Real-world rollouts expose failure modes that are partly hidden by binary success rates.

\paragraph{Model-level patterns.}
XSkill produces the most consistent real-world behavior among the evaluated models. On seen tasks, its rollouts usually preserve the demonstrated action sequence and tolerate moderate object-placement changes; on unseen tasks it also shows useful partial transfer, but the records still contain failures from insufficient lift, premature gripper closure, or failure to release at the target. ACT/DINOv2 and DP/DINOv2 are less uniform. Their successful rollouts often follow the correct high-level subtask order, but many failures come from contact timing: the gripper closes before reaching the object, does not descend far enough, or releases too early/too late. $\pi_{0.5}$ is the least stable in zero-shot real-world transfer. Its seen-task and fine-tuned runs can complete some tasks, but unseen zero-shot records repeatedly show arm jitter, early closure, and left-gripper non-closure, so its language-conditioned interface does not by itself give reliable physical transfer.

\paragraph{Task-level patterns.}
The clearest real-world failure cases are tasks where a small contact error destroys the rest of the rollout. \textsc{Stir Soup in Bowl} fails frequently because the spoon is thin and slips or is never fully captured. \textsc{Discard Food Waste} is difficult across models because it combines small/deformable food items, bimanual coordination, and precise release; many failures are left-gripper non-closure or no release after grasp. \textsc{Scan Beverage Barcode} similarly stresses grasp stability on the scanner or beverage, and several models reach the object but fail to close or hold the gripper. \textsc{Hang Cup on Rack} is a height-sensitive placement task: policies often grasp the mug but fail to lift high enough or align with the rack. By contrast, \textsc{Place Plates in Rack}, \textsc{File Medical Documents}, \textsc{Fold Towel into Basket}, and \textsc{Seal and Pack Box} show more coherent behavior, because the required grasps and placements are wider-tolerance and less dependent on exact gripper contact.

\paragraph{Unseen transfer.}
The unseen tasks should be read as evidence of partial generalization, not solved real-world imitation. \textsc{Return Remote to Box} is representative: several policies can localize and grasp the remote, and performance is often reasonable at L0--L2, but L3 exposes the missing affordance adaptation--the object is grasped with a plausible motion but placed at the wrong location, not lifted high enough, or released before reaching the substitute container. \textsc{Pour Water into Cup} also improves with adaptation, but failures still arise from missing the handle or closing before contact. The released rollout metadata stores task, level, model, corpus scale, setting, episode identifier, and video path, so each qualitative failure mode can be traced to representative rollouts in the benchmark release. These cases suggest that the models learn task-relevant visual and motion priors from IG-10K, while the few-shot real-world data are still too small to calibrate precise grasp points, release timing, and level-specific affordance substitutions.

\paragraph{Simulation per-task view.}
Table~\ref{tab:per-task-reference} gives the same diagnosis at larger scale. Tasks such as \textsc{FoldBox}, \textsc{GrindFood}, \textsc{PlaceMugRack}, and \textsc{PickRemoteControl} have high mean success, indicating that the policies can often infer the object and intended task. Yet the per-level breakdown is more informative than the mean: for \textsc{PickRemoteControl}, both DP and ACT remain strong through L0--L2, while L3 drops, matching rollouts where the remote can be found or grasped but the changed target relation is not executed correctly. Lower-scoring tasks such as \textsc{PlaceCupPlate}, \textsc{PlaceFoodScale}, and \textsc{PlaceCommodityRack} require tighter spatial alignment and contact geometry, so failures are better interpreted as insufficient manipulation coverage and calibration rather than a complete absence of task understanding.

\section{Per-Task Reference Results}
\label{app:per-task-reference}

\begingroup
\small
\setlength{\tabcolsep}{2.3pt}
\renewcommand{\arraystretch}{0.98}
\begin{longtable}{@{}>{\raggedright\arraybackslash}p{0.35\textwidth}rrrrrrrrrr@{}}
\caption{\textbf{Per-task reference results for simulation success rate.} Each cell reports success rate (\%) over 10 evaluation episodes for the corresponding task and imitation level.}
\label{tab:per-task-reference}\\
\toprule
\multirow{2}{*}{Task} & \multicolumn{4}{c}{DP} & \multicolumn{4}{c}{ACT} & \multicolumn{2}{c}{Mean} \\
\cmidrule(lr){2-5}\cmidrule(lr){6-9}\cmidrule(l){10-11}
& L0 & L1 & L2 & L3 & L0 & L1 & L2 & L3 & DP & ACT \\
\midrule
\endfirsthead
\caption[]{\textbf{Per-task reference results for simulation success rate} (continued).}\\
\toprule
\multirow{2}{*}{Task} & \multicolumn{4}{c}{DP} & \multicolumn{4}{c}{ACT} & \multicolumn{2}{c}{Mean} \\
\cmidrule(lr){2-5}\cmidrule(lr){6-9}\cmidrule(l){10-11}
& L0 & L1 & L2 & L3 & L0 & L1 & L2 & L3 & DP & ACT \\
\midrule
\endhead
\midrule
\multicolumn{11}{r}{\footnotesize Continued on next page}\\
\endfoot
\bottomrule
\endlastfoot
CleanCup & 40 & 50 & 20 & 100 & 30 & 30 & 70 & 90 & 52.5 & 55.0 \\
CleanDesk & 10 & 60 & 40 & 60 & 80 & 100 & 100 & 100 & 42.5 & 95.0 \\
CutFruit & 100 & 50 & 0 & 100 & 100 & 100 & 0 & 100 & 62.5 & 75.0 \\
FoldBox & 100 & 100 & 100 & 100 & 100 & 100 & 100 & 100 & 100.0 & 100.0 \\
FoldTowel & 60 & 40 & 10 & 40 & 100 & 100 & 100 & 80 & 37.5 & 95.0 \\
GrindFood & 100 & 100 & 100 & 100 & 90 & 100 & 100 & 70 & 100.0 & 90.0 \\
KnifeBowlFork & 0 & 60 & 60 & 90 & 100 & 100 & 80 & 50 & 52.5 & 82.5 \\
LiftLid\allowbreak FromSkillet & 0 & 80 & 80 & 100 & 100 & 100 & 100 & 100 & 65.0 & 100.0 \\
OpenBox & 100 & 90 & 0 & 100 & 100 & 100 & 0 & 100 & 72.5 & 75.0 \\
OpenLiquidCap & 60 & 50 & 70 & 50 & 90 & 60 & 90 & 0 & 57.5 & 60.0 \\
PickApple\allowbreak Banana\allowbreak ToBaskets & 50 & 100 & 50 & 100 & 90 & 90 & 50 & 50 & 75.0 & 70.0 \\
PickAppleBasket & 70 & 70 & 100 & 80 & 100 & 90 & 100 & 100 & 80.0 & 97.5 \\
PickApple\allowbreak ToScale & 40 & 90 & 40 & 80 & 90 & 10 & 90 & 80 & 62.5 & 67.5 \\
PickFood & 40 & 100 & 10 & 90 & 100 & 70 & 30 & 90 & 60.0 & 72.5 \\
PickFruits\allowbreak ToPlate & 90 & 90 & 100 & 50 & 20 & 40 & 100 & 0 & 82.5 & 40.0 \\
PickPill\allowbreak ToRegions & 40 & 60 & 50 & 10 & 90 & 90 & 70 & 80 & 40.0 & 82.5 \\
PickRemote\allowbreak Control & 100 & 100 & 100 & 40 & 100 & 100 & 100 & 80 & 85.0 & 95.0 \\
PickTennisBall\allowbreak GolfBall & 90 & 90 & 90 & 100 & 60 & 80 & 70 & 50 & 92.5 & 65.0 \\
PickWash & 70 & 100 & 30 & 80 & 90 & 100 & 60 & 90 & 70.0 & 85.0 \\
PlaceBook\allowbreak Bookcase & 100 & 100 & 90 & 90 & 100 & 80 & 100 & 30 & 95.0 & 77.5 \\
PlaceBrushRest & 90 & 70 & 100 & 100 & 80 & 100 & 100 & 100 & 90.0 & 95.0 \\
PlaceBurgerTray & 80 & 90 & 80 & 70 & 90 & 80 & 90 & 70 & 80.0 & 82.5 \\
PlaceChipsRack & 90 & 80 & 100 & 80 & 70 & 70 & 70 & 100 & 87.5 & 77.5 \\
PlaceCloth\allowbreak Basket & 100 & 100 & 80 & 10 & 100 & 100 & 90 & 90 & 72.5 & 95.0 \\
PlaceCommodity\allowbreak Rack & 40 & 50 & 30 & 30 & 100 & 100 & 90 & 0 & 37.5 & 72.5 \\
PlaceCupPlate & 40 & 20 & 70 & 30 & 50 & 60 & 70 & 0 & 40.0 & 45.0 \\
PlaceFileFolder & 80 & 70 & 100 & 80 & 80 & 80 & 60 & 100 & 82.5 & 80.0 \\
PlaceFoodScale & 40 & 20 & 90 & 0 & 60 & 100 & 100 & 0 & 37.5 & 65.0 \\
PlaceFruitBox & 80 & 20 & 90 & 30 & 70 & 30 & 90 & 60 & 55.0 & 62.5 \\
PlaceMagazine\allowbreak Folder & 80 & 100 & 0 & 70 & 100 & 100 & 0 & 70 & 62.5 & 67.5 \\
PlaceMugRack & 100 & 70 & 80 & 90 & 100 & 100 & 80 & 100 & 85.0 & 95.0 \\
PlacePillBox & 80 & 70 & 60 & 30 & 100 & 60 & 80 & 70 & 60.0 & 77.5 \\
PlacePlateRack & 90 & 40 & 30 & 80 & 100 & 90 & 50 & 80 & 60.0 & 80.0 \\
PlaceScrewdriver & 30 & 20 & 30 & 30 & 100 & 80 & 40 & 100 & 27.5 & 80.0 \\
PlaceShoeBox & 90 & 90 & 60 & 100 & 100 & 100 & 90 & 90 & 85.0 & 95.0 \\
PourCup & 10 & 100 & 100 & 90 & 100 & 100 & 100 & 100 & 75.0 & 100.0 \\
PourKetchup\allowbreak Fries & 50 & 50 & 100 & 30 & 100 & 100 & 80 & 60 & 57.5 & 85.0 \\
PourKettle & 90 & 80 & 0 & 80 & 100 & 90 & 0 & 100 & 62.5 & 72.5 \\
PourLiquidCup & 10 & 0 & 0 & 10 & 10 & 10 & 0 & 0 & 5.0 & 5.0 \\
PourLiquid\allowbreak Filter & 60 & 60 & 20 & 50 & 100 & 100 & 90 & 90 & 47.5 & 95.0 \\
PourLiquidMug & 60 & 10 & 100 & 20 & 0 & 90 & 70 & 0 & 47.5 & 40.0 \\
PressJuicer & 60 & 80 & 0 & 100 & 20 & 100 & 0 & 100 & 60.0 & 55.0 \\
PressStapler & 100 & 70 & 100 & 90 & 50 & 90 & 40 & 40 & 90.0 & 55.0 \\
PutBox & 100 & 90 & 90 & 10 & 100 & 90 & 100 & 100 & 72.5 & 97.5 \\
PutCube\allowbreak OnScale & 100 & 90 & 90 & 60 & 100 & 100 & 100 & 70 & 85.0 & 92.5 \\
ScanMilkBox & 100 & 100 & 100 & 90 & 100 & 100 & 100 & 100 & 97.5 & 100.0 \\
ScanPillBottle & 90 & 50 & 90 & 10 & 100 & 100 & 90 & 0 & 60.0 & 72.5 \\
StirSpoon & 20 & 0 & 40 & 10 & 50 & 50 & 40 & 80 & 17.5 & 55.0 \\
TransFood & 50 & 90 & 30 & 30 & 60 & 90 & 0 & 20 & 50.0 & 42.5 \\
WipePot & 10 & 0 & 0 & 100 & 20 & 0 & 0 & 80 & 27.5 & 25.0 \\
\end{longtable}
\endgroup

\end{document}